\pdfoutput=1

\documentclass[11pt]{article}
\usepackage{overpic}
\usepackage{float}

\newif\ifcomment\commenttrue

\usepackage[a-1b]{pdfx}

\usepackage{framed}
\usepackage{lmodern}
\usepackage{mdwlist}
\usepackage{siunitx}
\usepackage{latexsym}
\usepackage{colortbl}
\usepackage{xcolor}
\usepackage{nicefrac}
\usepackage{booktabs}
\usepackage{fnpct}
\usepackage{amsfonts}
\usepackage[T1]{fontenc}
\usepackage{bold-extra}
\usepackage{amsmath}
\usepackage{amssymb}
\usepackage{bm}
\usepackage{graphicx}
\usepackage{mathtools}
\usepackage{microtype}
\usepackage{multirow}
\usepackage{multicol}
\usepackage{xpatch}
\usepackage{latexsym,comment}
\usepackage[normalem]{ulem}

\newcommand*{\missingreference}{{\Huge \colorbox{red}{?reference?}}}
\newcommand*{\missingcitation}{{\Huge \colorbox{red}{?citation?}}}

\makeatletter
\xpatchcmd{\@setref}{\bfseries}{\missingreference}{}{}
\def\@citex[#1]#2{\leavevmode
    \let\@citea\@empty
    \@cite{\@for\@citeb:=#2\do
        {\@citea\def\@citea{,\penalty\@m\ }%
            \edef\@citeb{\expandafter\@firstofone\@citeb\@empty}%
            \if@filesw\immediate\write\@auxout{\string\citation{\@citeb}}\fi
            \@ifundefined{b@\@citeb}{\hbox{\reset@font\missingcitation}%
                \G@refundefinedtrue
                \@latex@warning
                {Citation `\@citeb' on page \thepage \space undefined}}%
            {\@cite@ofmt{\csname b@\@citeb\endcsname}}}}{#1}}
\makeatother

\usepackage{csquotes}
\newcommand{\quoted}[1]{\enquote{\textit{#1}}}

\newcommand{\mm}[0]{\textsc{llm}\xspace}

\newcommand{\gem}[1]{\mbox{\textsc{gem}}}
\newcommand{\abr}[1]{\textsc{#1}\xspace}

\newcommand{\hidetext}[1]{}
\newcommand{\ignore}[1]{}

\ifcomment
    \newcommand{\pinaforecomment}[3]{\colorbox{#1}{\parbox{.8\linewidth}{#2: #3}}}

    \newcommand{\prtodo}[1]{\pinaforecomment{lightblue}{pr}{#1}}
    \newcommand{\prtodoi}[1]{\pinaforecomment{lightblue}{pr}{#1}}
\else
    \newcommand{\pinaforecomment}[3]{}
    \newcommand{\prtodo}[1]{}
    \newcommand{\prtodoi}[1]{}
\fi

\newcommand{\smallurl}[1]{ \begin{tiny}\url{#1}\end{tiny}}

\definecolor{lightblue}{HTML}{3cc7ea}
\definecolor{CUgold}{HTML}{CFB87C}
\definecolor{grey}{rgb}{0.95,0.95,0.95}
\definecolor{ceil}{rgb}{0.57, 0.63, 0.81}
\definecolor{UMDred}{HTML}{ed1c24}
\definecolor{UMDyellow}{HTML}{ffc20e}

\newcommand{\nlp}[0]{\abr{nlp}}
\newcommand{\ai}[0]{\abr{ai}}

\usepackage{acl2023}

\usepackage{xspace}
\usepackage{xcolor}
\usepackage[utf8]{inputenc}
\usepackage{pgfplots}
\usepackage{dsfont}
\DeclareUnicodeCharacter{2212}{−}
\usepgfplotslibrary{groupplots,dateplot}
\usetikzlibrary{patterns,shapes.arrows}
\pgfplotsset{compat=newest}

\definecolor{myblue}{RGB}{30,90,200}

\usepackage{tikzscale}
\usepackage{relsize}
\usepackage{amsmath,amssymb}
\newcommand{\probP}{\text{I\kern-0.15em P}}

\definecolor{takeawaycolor}{HTML}{de0d0d}
\newcommand\takeaway[1]{\textcolor{takeawaycolor}{\textbf{#1}}}

\usepackage{multirow, colortbl}

\usepackage{tabularx,booktabs}
\usepackage{makecell}
\usepackage{subcaption}

\usepackage[normalem]{ulem}
\useunder{\uline}{\ul}{}

\usepackage{graphicx}

\definecolor{ablation6}{HTML}{fcefed}
\definecolor{ablation_tie}{HTML}{fce3e1}

\definecolor{ablation5}{HTML}{fcd8d4}
\definecolor{ablation4}{HTML}{FBC3BC}
\definecolor{ablation3}{HTML}{F7A399}
\definecolor{ablation2}{HTML}{F38375}
\definecolor{ablation1}{HTML}{EF6351}

\newcommand{\inlinecode}[1]{%
    \begin{tikzpicture}[baseline=0ex]%
         \node[anchor=base,%
         text height=0.7em,%
         text depth=0.7ex,%
         inner ysep=0pt,%
         draw=lightgray!50,%
         fill=lightgray!50,%
         rounded corners=2pt] at (0,0) {\footnotesize\texttt{#1}};%
    \end{tikzpicture}%
}

\definecolor{OliveGreen}{rgb}{0.05, 0.75, 0.24}
\definecolor{BrickRed}{rgb}{0.8, 0.25, 0.33}

\usepackage[]{algpseudocode}
\usepackage[]{algorithm}
\usepackage{float}
\algtext*{EndFor}%
\algtext*{EndProcedure}%

\usepackage{booktabs}
\usepackage[normalem]{ulem}
\useunder{\uline}{\ul}{}

\usepackage{times}
\usepackage{latexsym}
\usepackage{adjustbox}

\usepackage[T1]{fontenc}
\usepackage{amsmath}
\usepackage{amssymb}
\usepackage{booktabs}
\usepackage{tikzscale}
\usepackage{amsmath}

\usepackage{multirow, colortbl}

\usepackage{tabularx,booktabs}
\usepackage{makecell}
\usepackage{multirow}
\usepackage{scalerel,xparse}

\usepackage{cleveref}
\usepackage{microtype}
\usepackage[most]{tcolorbox}

\usepackage{enumitem}
\crefformat{section}{\S#2#1#3}
\crefformat{subsection}{\S#2#1#3}
\crefformat{subsubsection}{\S#2#1#3}

\definecolor{bggray}{rgb}{0.95, 0.95, 0.95}
\usepackage[%
    framemethod=tikz,
    skipbelow=\topskip,
    skipabove=\topskip
]{mdframed}
\mdfsetup{%
    leftmargin=0pt,
    rightmargin=0pt,
    backgroundcolor=bggray,
    middlelinecolor=black,
    roundcorner=3
}

\definecolor{SkyBlue}{rgb}{0.53, 0.81, 0.92}

\newtcolorbox[
  list inside=prompt,
  auto counter,
  number within=section
]{prompt}[1][]{%
  enhanced,
  float*=t,                 
  colbacktitle=black!60,
  fonttitle=\small,
  coltitle=white,
  fontupper=\footnotesize,
  boxsep=4pt,
  left=0pt, right=0pt, top=0pt, bottom=0pt,
  boxrule=1pt,
  width=\textwidth,          
  enlarge left by=0mm,
  enlarge right by=0mm,
  listing only,
  listing options={
    basicstyle=\ttfamily\footnotesize,
    breaklines=true,
    breakatwhitespace=true,
    language=json
  },
  #1,
}

\definecolor{circlered}{HTML}{fd6e51}
\definecolor{circleblue}{HTML}{2596be}
\definecolor{circleorange}{HTML}{f6b621}

\newtcolorbox[
  list inside=trace,
  auto counter,
  number within=section
]{trace}[1][]{%
  enhanced,
  float*=t,
  colback=blue!5,             
  colbacktitle=blue!60!black, 
  colframe=blue!60!black,     
  fonttitle=\small,
  coltitle=white,
  fontupper=\footnotesize,
  boxsep=4pt,
  left=0pt, right=0pt, top=0pt, bottom=0pt,
  boxrule=1pt,
  width=\textwidth,
  enlarge left by=0mm,
  enlarge right by=0mm,
  listing only,
  listing options={
    basicstyle=\ttfamily\footnotesize,
    breaklines=true,
    breakatwhitespace=true,
    language=json
  },
  #1,
}

\definecolor{UMDred}{HTML}{ed1c24}

\usepackage{color-edits}
\addauthor{vc}{blue}

\definecolor{yellowcolor}{HTML}{ffc20e}
\definecolor{redcolor}{HTML}{e99999}
\definecolor{orangecolor}{HTML}{f6b26b}
\definecolor{yellowcolor}{HTML}{ffd966}
\definecolor{bluecolor}{HTML}{a0c5e8}
\definecolor{purplecolor}{HTML}{d9d2e9}

\usepackage{tikz}

\usepackage{xcolor}
\definecolor{highlight}{HTML}{FFAE02}

\title{ 
(Im)Paired Programming:\\
Coding Agents Improve Productivity but Harm Understanding
}

\newcommand{\authorSpacing}{0.6cm}
\author{
\textbf{Nishant Balepur}$^{1, 2}$\hspace{\authorSpacing}
\textbf{Connor Baumler}$^{1}$ \hspace{\authorSpacing}
\textbf{Valerie Chen}$^{3}$ \hspace{\authorSpacing}\\
\textbf{Eunsol Choi}$^{2}$ \hspace{\authorSpacing}
\textbf{Rachel Rudinger}$^{1}$ \hspace{\authorSpacing}
\textbf{Jordan Boyd-Graber}$^{1}$ \\[0.5em]
$^{1}$University of Maryland \hspace{0.3cm}
$^{2}$New York University \hspace{0.3cm}
$^{3}$Carnegie Mellon University\\[0.5em]
\texttt{nbalepur@umd.edu} \hspace{0.5em} \texttt{jbg@umiacs.umd.edu}
}

\begin{document}
\maketitle

\begin{abstract}
Coding agents (e.g., Cursor) improve developer productivity by optimizing task completion, but shifting users from writing code to prompting and reviewing may harm their understanding---impeding oversight, learning, and communication.
To probe this, we have 54 students~create a website with one of two \ai{} systems:~an~\textit{agent} that edits user code; or a \textit{chatbot} where users write code alone or adapt generic code snippets.
We test understanding via comprehension questions and a task where users extend their code without agents, showing:
(1) While agents aid initial task completion, they harm users' code comprehension and thus~do not prepare users to extend their code;
(2) Low-effort agent interaction types---like copy+paste prompts and auto-accepted edits---are linked with lower comprehension; and
(3) Despite self-reported weaker understanding, users still prefer coding agents because they are quick and easy to use.
While users stay in the loop for coding workflows, understanding should not be forgotten.
Towards this goal, we distill our analyses into future~research directions for coding agent developers: dissuading low-effort prompting, creating readable code, and promoting active engagement.
\footnote{https://github.com/nbalepur/impaired-programming}
\end{abstract}

\section{Intro: Coding Under the Influence} \label{section:intro}

\begin{figure}
    \centering
    \includegraphics[width=\linewidth]{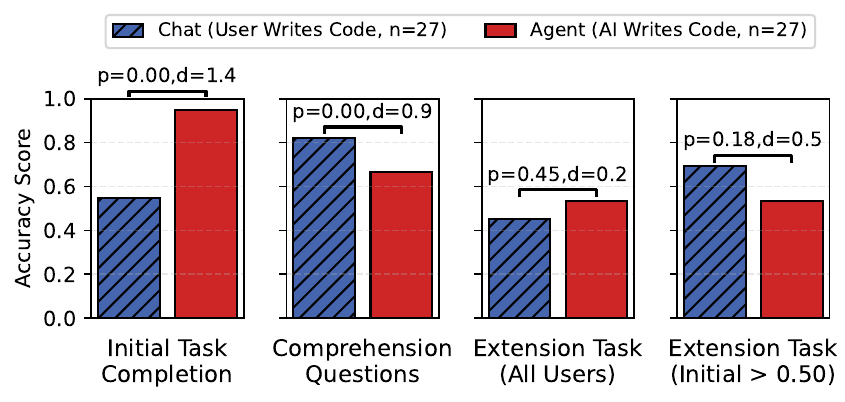}
    \vspace{-4.5ex}
    \caption{Our user study compares two~\ai{}s---an agent that writes code for users and a chatbot that~ensures our users write code---for finishing a web development task.
    Relative to chatbots, agents initially help users~complete the task but largely harm comprehension~\citep[$p<0.002$, Cohen's $d>0.80$]{cohen2013statistical}.
    Agent users can better extend their code without agents, but this~reverses when controlling for initial task accuracy, suggesting gains are driven by initial success over improved understanding.
    } \label{fig:teaser} 
    \vspace{-1ex}
\end{figure}
\begin{figure*}
    \centering
    \includegraphics[width=\linewidth]{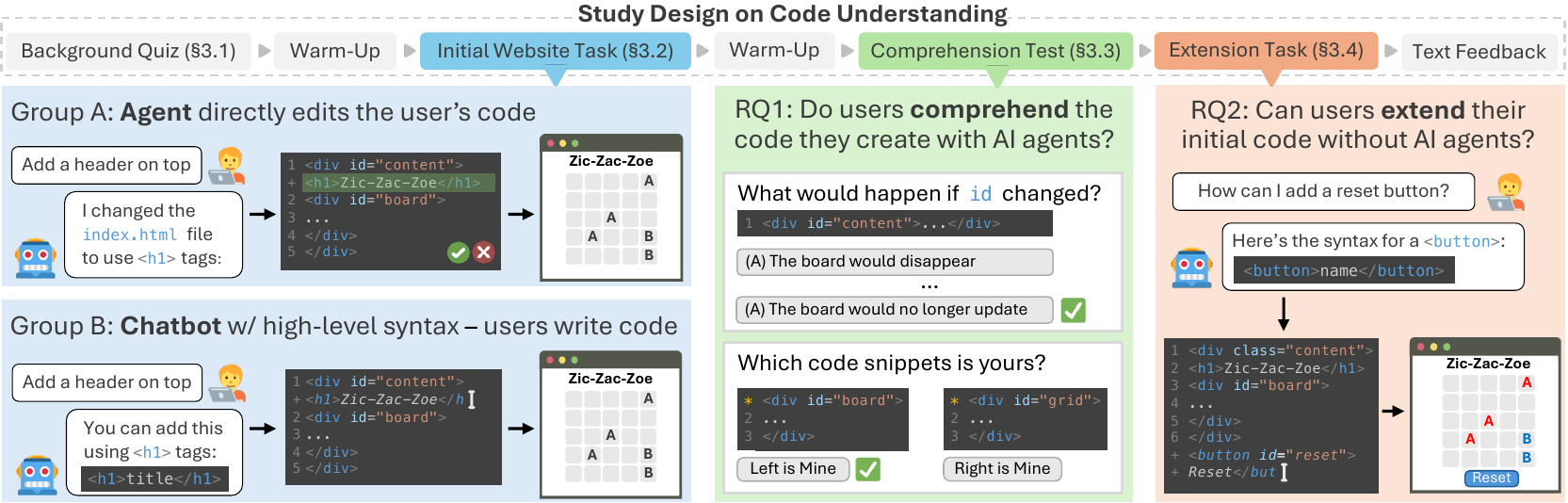}
    \vspace{-3.8ex}
    \caption{Our between-subjects user study to show how coding agents neglect user understanding. After taking a background quiz, $n=54$ users design a web-based variant of tic-tac-toe, prompting either: (A) an \ai{} agent~that directly edits user code; or (B) a chatbot that gives high-level syntax. We test understanding via~comprehension~questions about the user's code, and accuracy on a follow-up task where users must extend their code without the agent.}
    \label{fig:study}
    \vspace{-1ex}
\end{figure*}

Coding agents follow natural language queries~to modify code for users \citep[e.g., Cursor]{research2026composer}. They are~increasingly popular in~\nlp{},~evinced by the advent of ``Code Models'' and ``\mm{} Agents'' tracks at EMNLP.\footnote{https://2026.emnlp.org/calls/main\_conference\_papers/}
Researchers and developers~often equate agent progress to task completion metrics in benchmarks \citep{jimenez2024swebench}, then~validate users prefer agents and can use them to finish tasks quickly~\citep{peng2023impact, shao2026collaborative}.
By this standard, coding agents appear successful.

If coding agent progress aims to mirror real~use, task completion is not enough: programmers~often \textit{understand} their code---knowing its content and behavior~\citep{kam2025professional, qiao2026systematic}.~Understanding helps users stay productive if \ai{} is~unavailable~\citep[e.g., confidential work, outages]{liu2026ai}, explain their work for others to use \citep[e.g., research papers, hand-offs]{ko2007information}, and thwart catastrophic errors \citep[e.g., database deletion\footnote{Like this unfortunate case: https://tinyurl.com/claude-DB}]{engels2026scaling}.
Users who write~code alone often understand it~\citep{kolb2014experiential},~but~what happens when agents usurp writing and programming~turns into prompting and reviewing \citep{chen2026code}?


This paper tests whether users understand their code after working with coding agents optimized for task completion.
54 CS students make a website in a code editor with one of two prompt-based \ai{} tools: (1)~an~agent that directly edits the user's~code; and (2) a chatbot that forces users to actively write code from scratch or adapt generic code snippets (Figure~\ref{fig:study}, left). 
We then assess user understanding in two ways: comprehension questions tailored to each user's code (Figure~\ref{fig:study}, center) and accuracy~on a follow-up task where users extend their code to pass new test cases without agents (Figure~\ref{fig:study}, right). 

While agent users finish the initial task more~accurately, they score substantially lower in comprehension questions and not much higher in the extension~task (Figure~\ref{fig:teaser}, \cref{subsection:comparison}).
For~agent~users, novices submit code as accurate as experts but~have~worse comprehension (\cref{subsection:regression}), so traditional coding skills remain valuable.
Lastly, we reveal a tradeoff~to~explain similar extension accuracy (\cref{subsection:mediation}): agents create higher-quality code at first, but lower comprehension~makes the code harder to use.
Thus, we~argue coding agents today do not aid understanding.

We then qualitatively review user interactions~to help explain why agent users struggle to comprehend their code, informing new research directions towards agents that support understanding.
Users who rely on lower-effort prompt (copy+paste) and review (auto-accept) strategies have lower~comprehension (\cref{subsection:prompting}, \cref{subsection:interactions}), motivating techniques~for refusing ``lazy'' requests.
Users have higher comprehension when agents write~simpler code, so code readability metrics \citep{buse2009learning} could benefit agent training (\cref{subsection:outputs}).
Lastly, we study users' feedback to find opportunities in routing implementation tasks between users or \ai{}, improving control, and ensuring gains in understanding are clear (\cref{section:users}).

As agents tempt programmers to offload writing, we argue understanding must help shape~progress: task completion alone neglects understanding.
Toward user understanding of \ai{} code, we contribute:\\
\noindent \textbf{1)} An online between-subjects study to assess how coding agents influence the understanding of users.\\
\noindent \textbf{2)} Evidence that coding agents improve initial task accuracy, harm understanding, and fail to improve extension accuracy---despite users preferring them.\\
\noindent \textbf{3)} Studies of user-agent interaction modes and user feedback to inform the design of \ai{} coding agents.\\
\noindent \textbf{4)} A released dataset of 54 user websites, prompts, \ai{} traces, background ability, and understanding.

\section{Background and Related Work} \label{section:related_work}

To motivate our study, we discuss how agent evaluations neglect user understanding (\cref{subsection:rw_coding_agents})---a gap software engineering research can help fill (\cref{subsection:rw_understanding}).

\subsection{Evaluating Coding Agents} \label{subsection:rw_coding_agents}

Coding agents like Codex \citep{chen2021codex},~Copilot \citep{stratton2024introduction}, Cursor \citep{research2026composer},~and Claude Code \citep{anthropic2026agentic} are increasingly popular.
Unlike chatbots \citep{carreira2022pyo} and autocomplete \citep{10.1145/3368089.3417058}, agents enable workflows where~\ai{} writes most code, while the user's primary role is prompting and reviewing.

Most work develops coding agents to finish tasks \citep{jimenez2024swebench, merrill2026terminalbench}, leading researchers to ask~if this metric fully supports users in practice \citep{mozannar2024realhumaneval}.
Prior evaluations show offline metrics convert to online benefits for users' productivity \cite{peng2023impact, cui2026effects},\footnote{\citet{becker2025measuring} is an exception, but they recognize flaws such as selection bias in evaluation \citep{we-are-changing-our-developer-productivity-experiment-design}.} but add new risks in security \citep{wahed-etal-2025-mocha, wang-etal-2025-cve} and long-term maintainability \citep{Ottenhof2026HowDA, Huang2026MoreCLA}.

Evaluating code understanding with agents is~unexplored: a risk that impedes users' knowledge and use of their code \citep{wyrich202340}.
The~most similar work shows agents optimized~for task completion harm learning of new skills \citep{Kazemitabaar2023StudyingTE, shen2026ai}, but understanding is distinct: learning tests if users acquire generalizable skills, but understanding tests if users can reason about and work with their \textit{own}~code.

\subsection{Measuring Code Understanding} \label{subsection:rw_understanding}

To develop metrics for user understanding with coding agents in our study, we draw on software comprehension research \citep{brooks1983towards}---a field dedicated to how users~reason about code.
Prior work tests understanding by asking users to recall \citep{sheppard1979modern}, trace \citep{peitek2018look}, debug \citep{arunachalam1996cognitive}, and extend code snippets \citep{besker2020influence}, or self-reporting it \citep{boehm1976quantitative}.
We synthesize these metrics for our user study, using comprehension questions about users' code (\cref{subsection:comprehension}), accuracy on an extension task~(\cref{subsection:extension}), and self-reported user feedback (\cref{section:users}).

Most work studies code understanding outcomes after manipulating aspects of pre-defined snippets \citep{wyrich202340}, such as readability \citep{johnson2019empirical}, complexity \citep{peitek2021program}, and structure \citep{wiese2019linking}.
Studying if users understand their own code is rare \citep{dasgupta2010not}---as writing code is often assumed to aid understanding \citep{kolb2014experiential}---but is now at risk with agents~replacing writing.
Most similar in this field, \citet{10.1145/3742413.3789121} test if users feel ``ownership'' of their code with \ai{}.
We instead design a study to reveal agent users struggle to understand their code (Figure~\ref{fig:teaser}).

\begin{figure*}
    \centering
    \includegraphics[width=\linewidth]{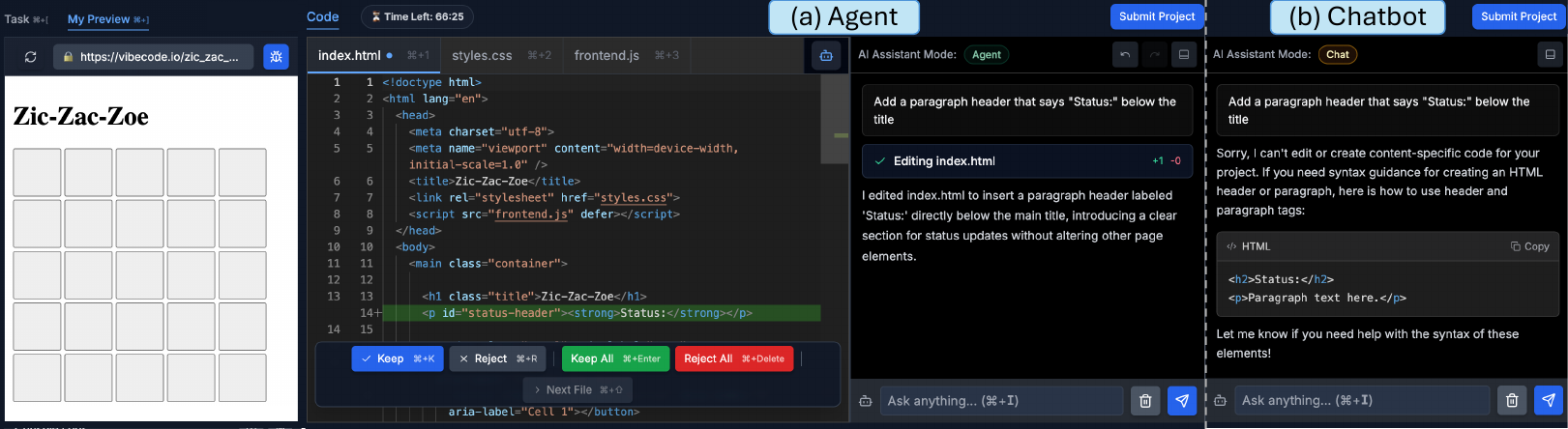}
    \vspace{-3.9ex}
    \caption{\small Overview of UI and \ai{} groups. Users create a website that meets input requirements (left) by prompting one of two \ai{} systems: (a) an agent that edits users code, explains edits, and affords user review; or (b) a chatbot that gives high-level syntax.}
    \label{fig:ui}
    \vspace{-1ex}
\end{figure*}

\section{Study Design: AI Website Development} \label{section:user_study}


Bridging research on agent evaluation and program comprehension (\cref{section:related_work}), we run a user study (Figure~\ref{fig:study}) to test productivity and understanding with coding agents.
We focus on web development, one of the most common coding agent use cases \citep{anthropic2025report}.
Users take a background quiz (\cref{subsection:background}), then we test productivity when users have agents write code versus mostly write code by themselves (\cref{subsection:initial}).
We finally instantiate code understanding metrics: users answer~comprehension questions about their code (\cref{subsection:comprehension}) and extend their initial code (\cref{subsection:extension}).
Below, we outline each step, then recruitment (\cref{subsection:recruitment}).

\subsection{Control for Analyses: Background Quiz} \label{subsection:background}

We first measure background~(BG) ability as a control for understanding~\citep{siegmund2015confounding}.
Since we study website~tasks (\cref{section:user_study}), we use multiple-choice HTML, CSS, and JavaScript questions from LinkedIn skill quizzes,\footnote{https://github.com/Ebazhanov/linkedin-skill-assessments-quizzes} popular for interview preparation.\footnote{Ideally, we would use a validated concept inventory \citep{10.1145/2839509.2844559}, but they do not exist for web development.}
We curate questions~to~target four skills in each programming language---syntax recall, conceptual knowledge, code tracing, and implementation \citep{Xie2019ATOA}---for 12 total items (Appendix~\ref{appendix:subsection:background}).
A front-end engineer validated questions.
Question accuracy measures BG ability.

\subsection{Productivity Metric: Initial Website Task} \label{subsection:initial}

We use one task across all users for controlled analyses, a standard user study design \citep{yan2024ivie, shen2026ai}.
We construct an original game creation task---a variation of tic-tac-toe called ``zic-zac-zoe''---where users modify three files for page structure (HTML), game logic (JavaScript), and style (CSS).
Thus, it is complex enough to test understanding while staying familiar.
We now outline the task, \ai{} groups, and productivity metrics.

\vspace{0.5em}

\noindent \textbf{Task Description:} In zic-zac-zoe, users start with a blank 5x5 board and must implement the following:
\begin{enumerate}[nosep]
\item Turns alternate between Player A/B, placing two symbols per turn (A$\to$A$\to$B$\to$B$\to...$).\footnote{Like tic-tac-toe \citep{beck1945combinatorial}, we believe zic-zac-zoe is unwinnable under optimal play, but still valid for our purposes.}
\item When a player clicks on an unoccupied square, it should display their symbol (``A'' or ``B'').
\item The game is over when a player occupies an entire row or column (not the diagonals), or no more moves can be played (i.e., a tie game).
\item The CSS should center the page horizontally.
\item A status element must show the next turn~(i.e., ``Player A Turn'' or ``Player B Turn'')~and a message when the game ends (e.g., ``A Wins!'').
\end{enumerate}

\noindent All users view criteria in text/video format and~have a 50 min. time limit, based on the time taken in~a~pilot study with three Ph.D. students.
Users can~read all criteria before starting and submit early if~done.

We only tell users to implement the above criteria; we do not mention that they will later answer comprehension questions or extend~their code.

\vspace{0.5em}

\noindent \textbf{\ai{} Groups:} We adapt VibeJam's~agentic coding UI \citep[Figure~\ref{fig:ui}]{balepur2026vibejam}, where users view the task description~and their website (left), write code in a VSCode-like IDE with HTML/CSS/JS starter files (middle), and prompt \ai{} for help (right).

To compare understanding when users actively write code or offload writing, we~randomly~assign users one of two prompt-based \ai{} using GPT-4.1:
\begin{enumerate}[nosep]
    \item Group A has an \textbf{agent} that directly edits~code (Figure~\ref{fig:ui}, a). We use the open-source~\textsc{Aider} agent; it executes prompts by proposing edits as diffs on users' code they can accept/reject, and summarizes executions~\citep{aider}.

    \item Group B has a \textbf{chatbot} that only gives generic code~snippets, so users must write code from scratch or manually adapt snippets (Figure~\ref{fig:ui}, b). This mirrors how programmers~adapt~code (e.g., StackOverflow) while keeping the same prompt interactions as the agent. It also~limits non-compliance rife in no-\ai{} studies \citep[e.g., using ChatGPT]{zhang2025generative}. The~chatbot is an \mm{} without access to users' code.~Its prompt rejects queries beyond syntax help and instructs~to only generate short text or generic code blocks of five or less lines (Prompt~\ref{prompt:chatbot}).

\end{enumerate}


\vspace{0.5em}

\noindent \textbf{Productivity Metric:}
We derive atomic rules from task criteria (e.g., ``\textit{The CSS should center the page horizontally}'' $\rightarrow$ ``\textit{The page is centered horizontally}'' and ``\textit{The page is centered via CSS}''), forming a 15-rule~rubric (Rubric~\ref{rubric:initial}).
We hide this rubric from users to limit gaming \citep{baker2008students}.
We score productivity as code accuracy---the proportion of met~rubric rules---and time-to-submission for secondary analyses.
Following \citet{sharma2026researchrubrics}, Gemini-3.1 Pro judges user code with our rubrics (Prompt~\ref{prompt:judge}).
To validate, Author A and B score ten agent and ten chatbot users' code via the rubric; they agree in 98\% of rules.
Gemini has near-perfect Cohen's $\kappa$ of $0.96$ with Author A~\citep{cohen1960coefficient}.


\begin{figure*}
    \centering
    \includegraphics[width=\linewidth]{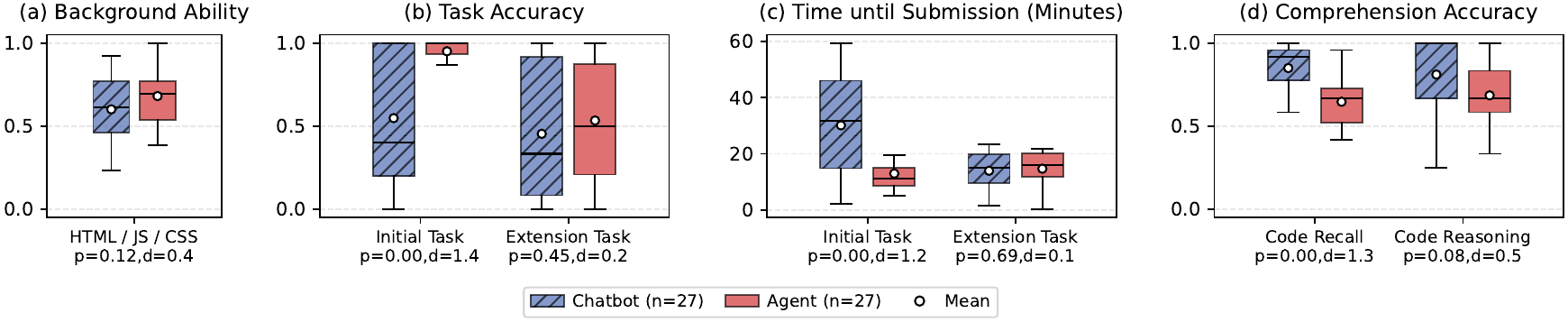}
    \vspace{-4ex}
    \caption{\small Metrics over \ai{} groups. Despite similar background ability (a), agent users are more accurate/quick in the initial task (b/c, left), but not when extending code~without agents (b/c, right) and agent users have worse comprehension (d).
    Appendix~\ref{appendix:subsection:outcomes} shows poor comprehension persists for 3/4 question types (all but \textit{change} questions). Coding agents do not support understanding.}
    \label{fig:boxplot}
    \vspace{-0.5ex}
\end{figure*}

\subsection{Understanding Metric 1: Comprehension} \label{subsection:comprehension}

Software engineering research often uses comprehension questions to assess understanding  \citep{wyrich202340}, but this is hard for code users write: you must ask about a user's code in real time.~Thus, our first understanding metric has \mm{}s tailor comprehension questions to users' code linked to two Bloom's taxonomy levels \citep{krathwohl2002revision}: 1)~if users can \underline{recall} elements in their code; and 2) how users \underline{reason} about their code behavior.
Users also self-report productivity, understanding, and preferences during zic-zac-zoe, which we study in~\cref{subsection:preferences}.

We use six questions of each type---two HTML, CSS, and JS---for 12 in total.
After the initial~task, users answer these \textit{without} seeing their code, apart from snippets we show.
Accuracy on comprehension~questions is our first understanding metric (Figure~\ref{fig:teaser}, middle).
We explain the question types next.

\vspace{0.4em}

\noindent \textbf{Recall:} We test how well users identify their~code \citep{sheppard1979modern}, predictive of memory/learning \citep{ebbinghaus_memory_1913, zindulka2026ai}.
We ask: (1) three multi-select questions where~users select which HTML elements, CSS selectors, and JS functions exist in their code~(Appendix Figure~\ref{fig:ui:recall_select}); and (2) three comparisons where users pick which of two logically-equivalent JS/HTML/CSS code snippets exist (Appendix Figure~\ref{fig:ui:recall_snippet}).
We randomly sample parts of users' code that they had to modify in the initial task for ground-truth answers and~use GPT-5.2 to create distractors (Prompts~\ref{prompt:distractor_identifier}, \ref{prompt:distractor_snippet}).

We validate 30 questions from pilot submissions for accuracy (e.g., answers exist in code).
To check for stylistic cues in \mm{} distractors, two~students answer 12 questions without seeing the code they are based on.
They achieve $0.38$ accuracy (random is $0.5$), so questions are not easily guessable from the choices alone \citep{lau2011guessing}.
Appendix~\ref{appendix:validate} validates this holds for real questions in our study.

\vspace{0.4em}

\noindent \textbf{Reasoning:} We design multiple-choice~questions to probe~how users reason about their code~\cite{peitek2018look}.
Each~question asks users about one of three main code snippets they had to write, each linked to a task criterion in \cref{subsection:initial}: HTML for status, CSS selector for page centering, and JS function for board rendering.\footnote{If the user did not attempt a specific criteria, we do not ask questions about it; GPT-5.2 checks for valid user attempts.} 
For each snippet, we ask two questions prior work suggests users can solve only if~they understand the code: (1) its \textit{purpose} \citep[Appendix Figure~\ref{fig:ui:reasoning}]{denny2024explaining}; and (2) how the website would \textit{change} if~the snippet changed in a given way  \citep[Appendix Figure~\ref{fig:ui:reasoning_change}]{zeller2009programs}.
To control for content and tailor questions to users, we design 12 question templates, then prompt GPT-5.2
to infill them using the user's code (Prompt~\ref{prompt:template}).

A~front-end engineer reviewed templates and~Author A ran five faulty and five perfect submissions to validate GPT aptly infills templates.~This holds for real questions during our study (Appendix~\ref{appendix:validate}).

\begin{figure*}
    \centering
    \includegraphics[width=\linewidth]{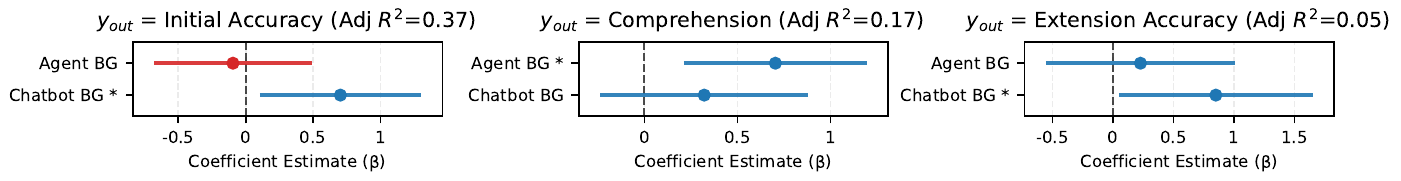}
    \vspace{-5ex}
    \caption{\small Regression coefficients with 95\% CIs when predicting study outcomes from user background (BG) ability, segmented by \ai{} groups. We show: (1) BG ability only predicts initial task accuracy in chatbot users, with agents equalizing users~regardless of BG; (2) BG strongly~predicts comprehension across groups; and (3) BG weakly predicts extension accuracy~(Adj. $R^2=0.05$).}
    \label{fig:regression}
    \vspace{-3ex}
\end{figure*}

\subsection{Understanding Metric 2: Extension Task} \label{subsection:extension}


We lastly test how well users edit their code without extensive \ai{} support, a downstream metric of code understanding~\citep{lucas2019does, nielebock2019commenting}.
Users start with their code submitted~in \cref{subsection:initial} and must extend it to pass the new criteria below:
\begin{enumerate}[nosep]
\item Add another win condition where players can win by occupying all four corners of the board.
\item Style the board symbols in the CSS file so that ``A'' symbols are red and ``B'' symbols are blue.
\item Add a ``Reset'' button to reset the entire game.
\end{enumerate}

\noindent All users finish this task with the chatbot. Like~\cref{subsection:initial}, no-\ai{} is difficult to enforce, but the chatbot still~requires users to actively write code---testing extension skill when users cannot offload writing to \ai{}.

We use the same \textsc{UI} as \cref{subsection:initial} and give 20 minutes, based on the time taken in pilot testing.
Agent users first finish a five minute warm-up task to familiarize with the chatbot.
We score task accuracy on a 12-rule~rubric (Rubric~\ref{rubric:extension}) via the same \mm{} judge in \cref{subsection:initial}.
After submission, we ask open-ended questions about their experiences, which we study~in~\cref{section:users}.

\subsection{User Recruitment} \label{subsection:recruitment}

We recruit 59 CS students in two U.S. colleges~(17 B.S., 31 M.S., 11 Ph.D.), as such~users~code with \ai{} \citep{vadaparty2024cs1}~and need to~understand their code in~coursework or research \citep{10.1145/3708359.3712104}.
Users report $5.4 \pm0.3$ years of coding experience in any language and $16.2 \pm1.4$ months coding with \ai{} (e.g., ChatGPT, Cursor).
We add attention checks in background (\cref{subsection:background}) and comprehension quizzes (\cref{subsection:comprehension}), and drop data of~five users who fail~them.
Users prompt $5.8 \pm 0.4$ (agent) and  $11.6 \pm 2.5$ (chatbot) times in the initial task, and  $6.8 \pm 0.9$ (agent) and $4.7 \pm 0.6$ (chatbot) times in the extension task.
Course instructors give extra~credit for taking the study, approved by \textsc{irb} (Ethics \cref{section:ethics}).

\section{Results: Outcomes with Coding Agents} \label{section:quant}

We now study our metrics to expose agent users are productive but lack understanding (\cref{subsection:comparison}).
We then show agents mask background skill (\cref{subsection:regression}) and give useful code for users initially, but users'~worse comprehension makes this code harder to extend~(\cref{subsection:mediation}).


\subsection{Coding Agents Harm Understanding} \label{subsection:comparison}

We first compare our productivity and understanding metrics across \ai{} groups.
Agents users~finish the initial task more~accurately/quickly than chatbot users, but these gains do not largely transfer to the extension task~(Figures~\ref{fig:teaser}, \ref{fig:boxplot}).
Agent users~score much lower on recall and reasoning questions, despite access to mechanisms meant for understanding: diff reviews and summaries of execution traces.
Overall, prompting and reviewing agent code did not support understanding as well as writing code \citep{shen2026ai}. 
We return~in \cref{section:qual} to study how to design \ai{} that improves user understanding.


\subsection{Background Skill Drives Comprehension} \label{subsection:regression}


We now test how user background (BG) influences understanding in our study.
We use linear~regression \citep{fisher1922goodness} to predict user outcomes $y_\text{out}$ (initial/extention accuracy, comprehension) via BG ability $x_\text{bg}$, \ai{} group $x_\text{group}$, and their~interaction:
\begin{equation}
    y_\text{out} \sim \beta_0 + \beta_{1} \cdot  x_{\text{group}} + \beta_2 \cdot x_{\text{bg}} + \beta_3 \cdot x_{\text{bg} \times \text{group}}, 
    \label{eq:regression_bg}
\end{equation}
where $x_{\text{group}}=1$ for chatbot and $0$ for agent users.
By setting $x_{\text{group}}=0$, we can use $\beta_2$ to capture how BG ability influences user outcomes with agents, and for $x_{\text{group}}=1$, we can use $\beta_2+\beta_3$ to capture how BG ability influences outcomes with chatbots. 

\begin{figure}
    \centering
    \includegraphics[width=\linewidth]{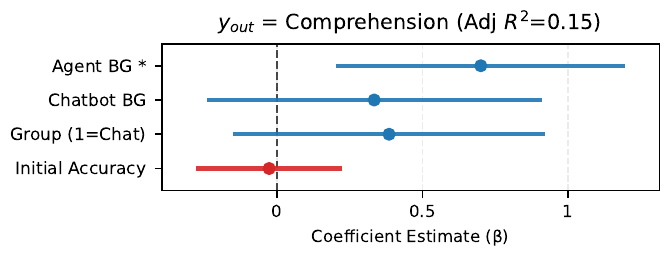}
    \vspace{-4.5ex}
    \caption{\small Regression coefficients with 95\% CIs after adding initial task accuracy to the comprehension regression (Equation~\ref{eq:regression_bg}).
    Initial accuracy barely improves adjusted $R^2$ ($0.40 \rightarrow 0.42$), so comprehension is distinct from task completion.}
    \label{fig:mediator1}
    \vspace{-1ex}
\end{figure}

Higher BG ability users have higher initial task scores with chatbots but not agents (Figure~\ref{fig:regression}, left)---all users score highly---so agent use can mask how evaluation settings (i.e., classrooms, hiring)~discern users with stronger BG \citep{xu2025ai}.
Comprehension scores positively predict BG ability (Figure~\ref{fig:regression}, middle), so traditional coding BG is valuable for understanding even with agents \citep{thorgeirsson2026computer}.
Lastly, higher BG predicts extension accuracy in chatbots but not agents (Figure~\ref{fig:regression}, right), so experts and novices may struggle similarly when extending their code without agents after agent use.


\begin{table*}[]
\small
\centering
\setlength{\tabcolsep}{2.1pt}
\begin{tabular}{@{}lcccc@{}}
\toprule
\textbf{Prompting Strategy}                                                                                 & Comp. ($\uparrow$) & BG Ability & \# Users & \probP(Used) \\ \midrule
Asking the agent to explain the codebase (``\textit{First, explain the whole project}'')                 & 0.635                        & 0.692      & 6 & 0.046        \\
Copying criteria (``\textit{Add a status element in a paragraph tag...}'')                  & 0.654                        & 0.668      & 25 & 0.793        \\
Iterative debugging (``\textit{Now there is a problem where it only shows player B...}'')  & 0.692                        & 0.662      & 5 & 0.040       \\
Turning criteria into syntax (``\textit{Broadcast \underline{statusMessage} to a \underline{p} tag under the \underline{h1} tag}'') & 0.743                        & 0.684      & 9 & 0.087        \\ \bottomrule
\end{tabular}
\vspace{-1.5ex}
\caption{\small Agent prompt strategies and the mean comprehension of the users who use them. Users who added code syntax to prompts had the best comprehension. Regressions support this trend (Appendix~\ref{appendix:interaction_regressions}). Appendix~\ref{appendix:chatbot_prompts} analyzes chatbot prompts.} \label{table:prompt}
\vspace{-2ex}
\end{table*}

\subsection{Agents Trade Accuracy for Understanding and this can Degrade Extension Ability} \label{subsection:mediation}


To learn what drives understanding, we add features to our regression (Eq.~\ref{eq:regression_bg}): 1) initial task accuracy to predict comprehension; and 2) initial task accuracy and comprehension to predict~extension accuracy.

Initial task accuracy has a small, insignificant coefficient for comprehension predictions (Figure~\ref{fig:mediator1}), showing that task completion and~comprehension are distinct optimization goals.
The regression also reveals that agent users still have significantly lower comprehension; thus, it is unlikely that this difference stems from \mm{}s producing easier~questions for the chatbot users who did not complete the task.

\begin{figure}
    \centering
    \includegraphics[width=\linewidth]{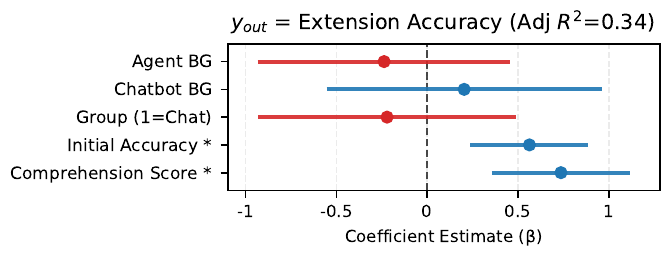}
    \vspace{-4.5ex}
    \caption{\small Regression coefficients with 95\% CIs after adding initial accuracy to the extension accuracy regression (Equation~\ref{eq:regression_bg}).
    Both feature coefficients are significantly positive~and largely boost $R^2$ ($0.05 \rightarrow 0.22$), showing why agent/chatbot users have similar extension accuracy: agents trade better~code scaffolds for worse comprehension, impeding code extension.}
    \label{fig:mediator2}
    \vspace{-0.5ex}
\end{figure}

Initial task accuracy and comprehension scores positively relate to extension accuracy (Figure~\ref{fig:mediator2}), with large adjusted $R^2$~gains from the initial model ($R^2 = 0.05 \rightarrow 0.34$).~This helps explain similar~extension accuracy over \ai{} groups (Figure~\ref{fig:teaser}): agents generate~higher-quality code scaffolds but sacrifice comprehension, impairing users' website extension when agents disappear \citep{collins1988cognitive}.
A~path mediation model confirms this in Appendix \ref{appendix:mediation}.

\section{Why Agent Users Lack Comprehension} \label{section:qual}

Agents harm comprehension (\cref{subsection:comparison})---which in turn may limit extension ability without agents (\cref{subsection:mediation})---so we now propose ways to design agents that aid comprehension.
We study steps of our agent workflow: prompting (\cref{subsection:prompting}) and reviewing (\cref{subsection:interactions}) agents, and attributes of the code (\cref{subsection:outputs})---concluding each with design takeaways for future work.
Our sample~size ($n=27$) matches \citet{shen2026ai} but is still moderate, so all analyses are exploratory.


\subsection{Discourage ``Lazy'' Prompting} \label{subsection:prompting}

Prompting affects learning \citep{srinath2025assessing},~so we~test how this shapes comprehension via qualitative coding~\citep{bingham2023data}.
Author A reviews~ten users' prompts and after~two~rounds, discovers four high-level strategies: directly copying task criteria, turning criteria into~code~syntax, iterative~debugging, and~eliciting explanations (Table~\ref{table:prompt}).
We then use Gemini-3 Flash to label a strategy for all $152$ prompts (Prompt~\ref{prompt:label}).
To validate, Author A and~B each label 50 random prompts, agreeing in 90\% of cases; Gemini and Author A agree in 94\% of~cases.

\takeaway{Takeaway 1:} Users who copied criteria had the lowest comprehension and users adding code~syntax had the highest, despite similar BG~(Table~\ref{table:prompt}).
This motivates the possibility for refusal \citep{bai2022training}: if~prompts use low-effort strategies (copying), agents could nudge users to review their code and rewrite their prompt in technical terms before proceeding.
This requires advances in classifiers to flag low-effort prompts \cite{song2007identifying} and post-training methods to imbue refusal in agents.

\begin{table}[]
\small
\centering
\renewcommand{\arraystretch}{0.8}
\setlength{\tabcolsep}{2.5pt}
\begin{tabular}{@{}lcccc@{}}
\toprule
\textbf{Interaction}                                                                                 & Comp. ($\uparrow$) & BG Ability & \# Users & \probP(Used) \\ \midrule
Auto-Accept                  & 0.615                        & 0.603      & 6 & 0.053        \\
Override Changes                 & 0.661                        & 0.682      & 15 & 0.318        \\ 
Hit Accept (All)   & 0.664                        & 0.669      & 20  & 0.492       \\
Hit Accept (Each)   & 0.777                        & 0.714      & 7  & 0.136       \\ \bottomrule
\end{tabular}
\vspace{-1.75ex}
\caption{\small Mean comprehension of users who engage in different \ai{} code review. Users who auto-accept agent changes have lower comprehension versus users who review each file.} \label{table:interaction}
\vspace{-1ex}
\end{table}

\subsection{Actively Engage Users During Review} \label{subsection:interactions}

Agent users mostly interact with their code by reviewing \ai{} edits \citep{10.1145/3771937}, but effort can vary; when agents edit code, users can:~1)~prompt again to auto-accept changes; 2) accept all changes by clicking ``Accept All''; 3) accept changes individually by clicking ``Accept'' per file; or 4) override the changes via manual edits or clicking ``Reject''.

\takeaway{Takeaway 2:} Agent users who review each file have higher comprehension than those who auto-accept edits (Table~\ref{table:interaction})---and~tend~to have higher BG scores---but~even these users do not reach chatbot users' mean comprehension (Figure~\ref{fig:teaser}).~For~comprehension, current reviewing interactions may never replace writing code \citep{kolb2014experiential}. Thus, \nlp{}+\textsc{hci} research has a gap in designing agent interactions that encourage active engagement versus passive review, boosting comprehension \citep{yan2025answering}.

\subsection{Generate More Readable Code} \label{subsection:outputs}

We study how the type of code agents create relates to comprehension.
We focus~on code readability---known to influence comprehension \citep{9240710}---and analyze agent users' final JavaScript file, as it hosts most of zic-zac-zoe's logic (\cref{subsection:initial}). 

We adopt standard readability metrics from~\citet{10.1145/1985441.1985454}: \textbf{1) lines of code}; \textbf{2) entropy}---how evenly tokens\footnote{We use the Esprima tokenizer: https://esprima.org/. Appendix~\ref{appendix:code_complexity} has implementation details and metric formulas.} (e.g., variables) are distributed (high entropy has more variety); and~\textbf{3)~volume}---total information~via unique token count and (1).
As documentation shapes comprehension \citep{brooks1982theoretical}, we also measure the \textbf{Prop}ortion of lines with \textbf{Comments}. We fit four regressions---predicting comprehension from each readability metric---as metrics have high colinearity \citep{10.1145/1985441.1985454}.

\begin{figure}
    \vspace{-1ex}
    \centering
    \includegraphics[width=\linewidth]{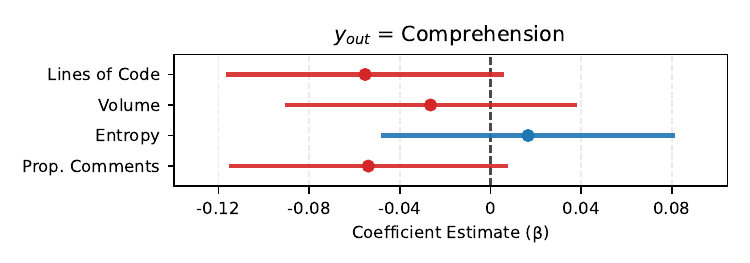}
    \vspace{-4.5ex}
    \caption{\small Model coefficients and 95\% CIs when using z-score normalized code readability metrics to predict agent user comprehension, fitting four regressions. Users with fewer lines of code and proportion of comments have better~comprehension.}
    \label{fig:complexity}
    \vspace{-1ex}
\end{figure}

\takeaway{Takeaway 3:} Users with JS code of~fewer lines, comments, and volume had higher comprehension (Figure~\ref{fig:complexity}), suggesting agent users better digest concise code~without distracting comments---observed in past \ai{} coding interviews~\citep{Barke2022GroundedCH}.
As code~properties~may~shape understanding, this~motivates optimizing agents to follow best readability practices \citep{maalej2014comprehension} beyond just task completion \citep{le2022coderl}.
This spurs challenges in using these metrics for training \citep{jain2025multiturn} while curbing reward hacking \citep{zhong2026impossiblebench}.

\section{How Users Perceive Coding Agents} \label{section:users}

To complement our analyses of observed user behavior (\cref{section:qual}), we now examine users' self-reported (\cref{subsection:preferences}) and written feedback (\cref{subsection:written_feedback}) to derive more directions for future \nlp{} research in coding agents.

\subsection{Users Recognize Reduced Understanding with Coding Agents, but Still Prefer Them} \label{subsection:preferences}

We review users' self-reported productivity, understanding, and preferences from~\cref{subsection:comprehension}.
Users find the agent more helpful and simpler than chatbots, but recognize weaker understanding (Table~\ref{table:ratings}).
Despite this, users prefer the agent over chatbots---but most prefer having the option to switch between the \ai{}s.

\takeaway{Takeaway 4:} Users may prefer \ai{} that optimizes for task completion, so improving understanding cannot rely solely on promoting alternative systems like chatbots.
However, as users prefer switching between \ai{}, future studies can assess whether~they do so responsibly, or design routers to predict when \ai{} or users should implement code to balance productivity and understanding \citep{miranda-etal-2025-hybrid}.

\begin{table}[!t]
\small
\centering

\setlength{\tabcolsep}{3.3pt}

\newcommand{\err}[2]{%
  \ensuremath{%
    \text{#1}\,\text{\scriptsize$\pm${#2}}%
  }%
}

\newcommand{\cellhl}[2]{\colorbox{#1}{\strut #2}}
\newcommand{\hlcaption}[2]{\raisebox{0pt}[0pt][0pt]{\colorbox{#1}{\strut #2}}}

\renewcommand{\arraystretch}{0.6}

\begin{tabular}{@{}lcc@{}}
\toprule
From 1--5, how much do you agree:          & Agent & Chatbot \\ \midrule

\cellhl{blue!20}{The agent/chatbot was helpful} & \err{\textbf{4.7}}{0.2} & \err{3.3}{0.2} \\
\cellhl{blue!20}{The agent/chatbot took mental effort} & \err{1.9}{0.2} & \err{\textbf{3.7}}{0.2} \\ \midrule

\cellhl{green!20}{I reviewed agent/chatbot responses}     & \err{3.2}{0.3} & \err{\textbf{4.0}}{0.2} \\
\cellhl{green!20}{The code I submitted feels like my own}      & \err{2.5}{0.2} & \err{\textbf{3.7}}{0.2} \\
\cellhl{green!20}{I understand how my code works}  & \err{3.3}{0.3} & \err{\textbf{4.2}}{0.2} \\
\cellhl{green!20}{I could easily extend my code}   & \err{3.1}{0.3} & \err{\textbf{3.2}}{0.2} \\ \midrule

\cellhl{orange!20}{I prefer no agent/chatbot in new tasks}        & \err{1.4}{0.2} & \err{1.9}{0.2} \\
\cellhl{orange!20}{I prefer the chatbot in new tasks}  & \err{3.2}{0.3} & \err{2.7}{0.2} \\
\cellhl{orange!20}{I prefer the agent in new tasks}    & \err{4.2}{0.2} & \err{3.9}{0.2} \\
\cellhl{orange!20}{I prefer switching between agent/chatbot}   & \err{\textbf{4.6}}{0.2} & \err{\textbf{4.1}}{0.3}\\

\bottomrule
\end{tabular}
\vspace{-1.5ex}
\caption{\small User 1--5 ratings on \hlcaption{blue!20}{usefulness}, \hlcaption{green!20}{understanding}, and \hlcaption{orange!20}{preferences} across agent and chatbot groups.
The group with the highest usefulness and understanding are \textbf{bold}, and the most favored \ai{} is \textbf{bold}.
Users report~the~agent as more helpful and easier to use than the chatbot but recognize it harms understanding. Still, users prefer using the agent. Appendix \ref{appendix:perception} shows these perceptions correlate with true understanding.
\label{table:ratings}}
\vspace{-3ex}
\end{table}

\subsection{Why Users Feel They Lack Understanding} \label{subsection:written_feedback}

Lastly, to learn why our users believe coding agents impair understanding, we review their written feedback from \cref{subsection:extension}, distilled into four common themes.

\vspace{0.5em}

\noindent \textbf{Who Needs to Think.} Coding with agents required little mental effort.
Many users conveyed \quoted{I didn't have to think at all.} (U10), especially since they did not need to use website or CS knowledge (U15, U49) and the agent had context from the users' files (U31), allowing users to prompt \quoted{fuzzier instructions} (U35).
As users could preview their website, they felt no desire to review changes (U27, U54).
U46 rarely reviewed, despite knowing it~would be \quoted{detrimental in the long run}, showing the need for lightweight interactions to promote  active thinking.



\vspace{0.5em}

\noindent \textbf{Agents Can Get Out of Control.} Guiding agent behavior was hard. U16 said agents would \quoted{change somewhere unexpected or even destroy the original~functions}, impairing understanding.
In U3's submission, the agent eagerly included features not needed in the task (e.g., board styling), further obscuring the code.
Future work can explore methods to add user control to execution \citep{feng2026cocoa} and ensure the execution trace summaries faithfully explain all changes agents make \citep{li-etal-2025-towards-better}.

\vspace{0.5em}

\noindent \textbf{Make Code Quality Actionable.} Users found~extension tasks easy when agents wrote high-quality code, citing comments (U44, U54), modular structure (U5, U46), and well-named functions (U21)---showing the benefits of rewarding these signals beyond task completion~\citep{jimenez2024swebench}.
However, these benefits resembled onboarding versus better understanding: \quoted{Comments made it easy to understand where to add new features, but having not fully reviewed the code made it much harder to understand everything} (U31).
Thus, code quality alone is insufficient; agents will need to guide users through output code to truly support understanding.

\vspace{0.5em}

\noindent \textbf{Understanding is Polarizing.} While many professional developers value understanding in \ai{} coding \citep{kam2025professional}, users were split.
U57 wished the agent \quoted{made me understand where code is supposed to go}, but most praised agents' accuracy and speed (U5, U26, U54).
For chatbot~users, U28 felt it \quoted{really helped remember how to use [the code] later on}, but others rejected it: U5 said \quoted{It's totally useless} and U49 noted its value for learning, but asked: \quoted{What's the point of learning all this when the AI can build it out for you?}
A core challenge in agent design is making benefits of understanding explicit, rather than assuming users will value it.






 
\section{Conclusion: What's Next for \ai{} Agents} \label{section:conclusion}

Task completion defines progress for coding agents \citep{shen2025completion}, but this goal alone erodes user understanding.
While we expose this empirically in our zic-zac-zoe website task, work remains in generalizing our understanding study across programming tasks like algorithmic design \citep{cormen2022introduction} and data analysis \citep{chambers1998programming}, testing longitudinal effects via comprehension quizzes every $n$ projects \citep{kosmyna2025your}, and whether existing interactions for coding agents \citep[e.g., plan mode, switching \ai{} modes]{feng2026cocoa} mitigate these issues.
Our open-sourced~study design, interface, and data will support these future efforts.

Beyond diagnosis, our future work aims to teach agents to support understanding.
Our analyses motivate jointly optimizing \ai{} for task completion and understanding signals based on code trace data \citep{chi2026editbench}, improving faithfulness in execution summaries \citep{zhang-etal-2023-extractive-summarization}, generating readable code \citep{jin2026reveal}, and more engaging \ai{} review mechanisms~\citep{sharma-etal-2024-investigating}.
Agents might soon automate many coding tasks, but if we want users to work with their code---monitoring for failures, writing research papers, and learning new skills---agents cannot neglect user understanding.

\section{Limitations} \label{section:limitations}

Our study is limited in population and domain:~we evaluate CS students on two website development tasks.
While common in \nlp{}+\textsc{hci} user studies with coding agents \citep{yan2024ivie, shen2026ai}, these findings may not apply to other user groups and tasks.
Thus, we encourage future work to extend our protocol to other domains; we open-source all our data and \textsc{ui} to facilitate these efforts.

Further, developers typically have custom workflows in practice \citep{chen2026code} but for feasibility and to control agent use, our interface~supports fewer workflow customization features.
Future research could test whether our results hold for users working with coding agents they choose---such as Cursor \citep{research2026composer} or Codex \citep{chen2021codex}---by using our \mm{}-tailored comprehension questions with proctored sessions \citep{thorgeirsson2026computer} or IDE plugins \citep{yan2024ivie}.

To ease the burden of manually scoring code and creating questions per user, we rely on imperfect \nlp{} tools.
Despite our checks---discussions~with a front-end engineer, manual review of all generated questions, and \mm{} validation with author-labeled websites---errors still may have slipped through;
in Appendix~\ref{appendix:validate}, we more rigorously analyze \mm{}-generated questions and do not detect major issues.
Future research can explore standardizing and rigorously evaluating \mm{}s to support user study design and analysis, such as qualitative coding tools \citep{lam2024concept}, question generators \citep{anugraha2026sparkme}, and rubric-based judges \citep{kim2023prometheus}.

Finally, while we argue that user understanding is traditionally a desired objective for programming agents in real-world workflows, we acknowledge there are cases that short-term productivity may be sufficient, such as personal projects just for fun \citep[e.g., vibe coding a personal website]{hu2026human}.
However, many novices attempt these projects to gain realistic programming experience---promoting learning, ownership, and engagement \citep{zhang2023study}.
As a result, coding agents that jointly improve productivity and understanding would still help these users---and should at least be an option.

\section{Ethical Considerations} \label{section:ethics}

Our study shows that coding agents allows users to finish tasks without comprehending the code~they write.
If this persists over longer periods, this~could impede and possibly weaken user's skill formation \citep{shen2026ai}.
Future work should consider these risks when designing coding agents, and we outline many new directions so \nlp{} researchers can design tools that encourage user understanding.

For our specific user study, we do not collect and release PII in our data and compensate our students with coursework credit.
To ensure students are not coerced into participating in our study, we provide an additional assignment that users can complete for the same extra credit---implementing a research paper in Python---that requires the same amount of time ($\sim2$ hours) as our study. 
This compensation structure was approved by our organization's \textsc{irb}.

We used Generative AI (GenAI) in this project.
We used Cursor\footnote{https://cursor.com/agents} to design our interface and plots, and ChatGPT to refine paper writing.
GenAI did not directly write any parts of this paper.
We take responsibility for GenAI errors.
By discussing~\ai{}~use here, we encourage \nlp{} researchers to do the same.

\section*{Acknowledgments}

We thank the \abr{clip} lab at the University of Maryland and the \abr{cilvr} lab at New York University for their support.
We are immensely grateful to Grace Chen for reviewing and validating our study questions, and Navita Goyal for extensive discussions on our modeling.
We thank Yu Hou, Dang Nguyen, Navita Goyal, and Paiheng Xu for pilot testing our interface.
We also appreciate discussions on earlier drafts from Vishakh Padmakumar, Wichayaporn Wongkamjan, Yu Hou, Deniz Qian, Fumeng Yang, David Weintrop, Shi Feng, Majeed Kazemitabaar, Aakanksha Naik, Joseph Chee Chang, and Pao Siangliulue.
This material is based upon work supported by the National Science Foundation under \abr{iis}-2339746 (Rudinger) \abr{iis}-2403436 (Boyd-Graber), and \abr{dge}-2236417 (Balepur).
Any opinions, findings, and conclusions or recommendations expressed in this material are those of the author(s) and do not necessarily reflect the views of the National Science Foundation.


\bibliography{custom}

@inproceedings{feng2026cocoa,
author = {Feng, K. J. Kevin and Pu, Kevin and Latzke, Matt and August, Tal and Siangliulue, Pao and Bragg, Jonathan and Weld, Daniel S and Zhang, Amy X. and Chang, Joseph Chee},
title = {Cocoa: Co-Planning and Co-Execution with AI Agents},
year = {2026},
isbn = {9798400722783},
publisher = {Association for Computing Machinery},
address = {New York, NY, USA},
url = {https://doi.org/10.1145/3772318.3791673},
doi = {10.1145/3772318.3791673},
booktitle = {Proceedings of the 2026 CHI Conference on Human Factors in Computing Systems},
articleno = {16},
numpages = {23},
location = {
},
series = {CHI '26}
}

@book{kolb2014experiential,
  title={Experiential learning: Experience as the source of learning and development},
  author={Kolb, David A},
  year={1984},
}

@inproceedings{
sharma2026researchrubrics,
title={ResearchRubrics: A Benchmark of Prompts and Rubrics For Evaluating Deep Research Agents},
author={Manasi Sharma and Chen Bo Calvin Zhang and Chaithanya Bandi and Clinton Wang and Ankit Aich and Huy Nghiem and Tahseen Rabbani and Ye Htet and Brian Jang and Sumana Basu and Aishwarya Balwani and Denis Peskoff and Marcos Ayestaran and Sean M. Hendryx and Brad Kenstler and Bing Liu},
booktitle={The Fourteenth International Conference on Learning Representations},
year={2026},
url={https://openreview.net/forum?id=ErnvfmSX0P}
}

@article{boehm1976quantitative,
  title={Quantitative evaluation of software quality},
  author={Boehm, Barry W and Brown, JR and Lipow, M},
  journal={Software Engineering: Barry W. Boehm's Lifetime Contributions to Software Development, Management, and Research},
  pages={25},
  year={1976},
  publisher={IEEE San Francisco, CA}
}

@misc{anthropic2025report,
  title        = {Anthropic Economic Index: AI’s impact on software development},
  author       = {{Anthropic}},
  institution  = {Anthropic},
  year         = {2025},
  month        = apr,
  url          = {https://resources.anthropic.com/hubfs/2026%20Agentic%20Coding%20Trends%20Report.pdf?hsLang=en}
}

@inproceedings{zindulka2026ai,
author = {Zindulka, Tim and Goller, Sven and Fernandes, Daniela and Welsch, Robin and Buschek, Daniel},
title = {The AI Memory Gap: Users Misremember What They Created With AI or Without},
year = {2026},
isbn = {9798400722783},
publisher = {Association for Computing Machinery},
address = {New York, NY, USA},
url = {https://doi.org/10.1145/3772318.3791494},
doi = {10.1145/3772318.3791494},
booktitle = {Proceedings of the 2026 CHI Conference on Human Factors in Computing Systems},
articleno = {61},
numpages = {22},
location = {
},
series = {CHI '26}
}

@inproceedings{lucas2019does,
  title={Does the introduction of lambda expressions improve the comprehension of Java programs?},
  author={Lucas, Walter and Bonif{\'a}cio, Rodrigo and Canedo, Edna Dias and Marc{\'\i}lio, Diego and Lima, Fernanda},
  booktitle={Proceedings of the XXXIII Brazilian symposium on software engineering},
  pages={187--196},
  year={2019}
}

@article{nielebock2019commenting,
  title={Commenting source code: is it worth it for small programming tasks?},
  author={Nielebock, Sebastian and Krolikowski, Dariusz and Kr{\"u}ger, Jacob and Leich, Thomas and Ortmeier, Frank},
  journal={Empirical Software Engineering},
  volume={24},
  number={3},
  pages={1418--1457},
  year={2019},
  publisher={Springer}
}

@article{beck1945combinatorial,
  title={Combinatorial games},
  author={Beck, J{\'o}zsef},
  journal={American history},
  volume={1861},
  number={1900},
  year={1945}
}

@misc{anthropic2026agentic,
  title        = {2026 Agentic Coding Trends Report: How Coding Agents Are Reshaping Software Development},
  author       = {{Anthropic}},
  institution  = {Anthropic},
  year         = {2026},
  month        = jan,
  url          = {https://resources.anthropic.com/hubfs/2026%20Agentic%20Coding%20Trends%20Report.pdf?hsLang=en}
}

@article{zhang2025generative,
  title={Generative AI meets open-ended survey responses: Research participant use of AI and homogenization},
  author={Zhang, Simone and Xu, Janet and Alvero, AJ},
  journal={Sociological Methods \& Research},
  volume={54},
  number={3},
  pages={1197--1242},
  year={2025},
  publisher={SAGE Publications Sage CA: Los Angeles, CA}
}

@article{Xie2019ATOA,
  title={A theory of instruction for introductory programming skills},
  author={Benjamin Xie and D. Loksa and Greg L. Nelson and Matthew J. Davidson and Dongsheng Dong and Harrison Kwik and A. H. Tan and Leanne Hwa and Min Li and Amy J. Ko},
  journal={Computer Science Education},
  year={2019},
  volume={29},
  pages={205 - 253},
  url={https://doi.org/10.1080/08993408.2019.1565235}
}

@article{qiao2026systematic,
  title={A systematic literature review of the use of GenAI assistants for code comprehension: Implications for computing education research and practice},
  author={Qiao, Yunhan and Shihab, Md Istiak Hossain and Hundhausen, Christopher},
  journal={ACM Transactions on Computing Education},
  volume={26},
  number={2},
  pages={1--33},
  year={2026},
  publisher={ACM New York, NY}
}

@software{balepur2026vibejam,
  author = {Nishant Balepur and Connor Baumler and Valerie Chen and Eunsol Choi and Rachel Rudinger and Jordan Lee Boyd-Graber},
  title = {VibeJam: An Open Platform for User Studies on Agentic Vibe Coding},
  year = {2026},
  publisher = {GitHub},
  url = {https://github.com/nbalepur/vibe-jam},
}

@article{maalej2014comprehension,
  title={On the comprehension of program comprehension},
  author={Maalej, Walid and Tiarks, Rebecca and Roehm, Tobias and Koschke, Rainer},
  journal={ACM Transactions on Software Engineering and Methodology (TOSEM)},
  volume={23},
  number={4},
  pages={1--37},
  year={2014},
  publisher={ACM New York, NY, USA}
}

@misc{peng2023impact,
  title={The Impact of AI on Developer Productivity: Evidence from GitHub Copilot}, 
      author={Sida Peng and Eirini Kalliamvakou and Peter Cihon and Mert Demirer},
      year={2023},
      eprint={2302.06590},
      archivePrefix={arXiv},
      primaryClass={cs.SE},
      url={https://arxiv.org/abs/2302.06590}, 
}

@inproceedings{kam2025professional,
  title={What do professional software developers need to know to succeed in an age of Artificial Intelligence?},
  author={Kam, Matthew and Miller, Cody and Wang, Miaoxin and Tidwell, Abey and Lee, Irene A and Malyn-Smith, Joyce and Perret, Beatriz and Tiwari, Vikram and Kenitzer, Joshua and Macvean, Andrew and others},
  booktitle={Proceedings of the 33rd ACM International Conference on the Foundations of Software Engineering},
  pages={947--958},
  year={2025}
}

@misc{research2026composer,
      title={Composer 2 Technical Report}, 
      author={Aaron Chan and Ahmed Shalaby and Alexander Wettig and Aman Sanger and Andrew Zhai and Anurag Ajay and Ashvin Nair and Charlie Snell and Chen Lu and Chen Shen and Emily Jia and Federico Cassano and Hanpeng Liu and Haoyu Chen and Henry Wildermuth and Jacob Jackson and Janet Li and Jediah Katz and Jiajun Yao and Joey Hejna and Josh Warner and Julius Vering and Kevin Frans and Lee Danilek and Less Wright and Lujing Cen and Luke Melas-Kyriazi and Michael Truell and Michiel de Jong and Naman Jain and Nate Schmidt and Nathan Wang and Niklas Muennighoff and Oleg Rybkin and Paul Loh and Phillip Kravtsov and Rishabh Yadav and Sahil Shah and Sam Kottler and Alexander M Rush and Shengtong Zhang and Shomil Jain and Sriram Sankar and Stefan Heule and Stuart H. Sul and Sualeh Asif and Victor Rong and Wanqi Zhu and William Lin and Yuchen Wu and Yuri Volkov and Yury Zemlyanskiy and Zack Holbrook and Zhiyuan Zhang},
      year={2026},
      eprint={2603.24477},
      archivePrefix={arXiv},
      primaryClass={cs.SE},
      url={https://arxiv.org/abs/2603.24477}, 
}

@inproceedings{
merrill2026terminalbench,
title={Terminal-Bench: Benchmarking Agents on Hard, Realistic Tasks in Command Line Interfaces},
author={Mike A Merrill and Alexander Glenn Shaw and Nicholas Carlini and Boxuan Li and Harsh Raj and Ivan Bercovich and Lin Shi and Jeong Yeon Shin and Thomas Walshe and E. Kelly Buchanan and Junhong Shen and Guanghao Ye and Haowei Lin and Jason Poulos and Maoyu Wang and Marianna Nezhurina and Di Lu and Orfeas Menis Mastromichalakis and Zhiwei Xu and Zizhao Chen and Yue Liu and Robert Zhang and Leon Liangyu Chen and Anurag Kashyap and Jan-Lucas Uslu and Jeffrey Li and Jianbo Wu and Minghao Yan and Song Bian and Vedang Sharma and Ke Sun and Steven Dillmann and Akshay Anand and Andrew Lanpouthakoun and Bardia Koopah and Changran Hu and Etash Kumar Guha and Gabriel H. S. Dreiman and Jiacheng Zhu and Karl Krauth and Li Zhong and Niklas Muennighoff and Robert Kwesi Amanfu and Shangyin Tan and Shreyas Pimpalgaonkar and Tushar Aggarwal and Xiangning Lin and Xin Lan and Xuandong Zhao and Yiqing Liang and Yuanli Wang and Zilong Wang and Changzhi Zhou and David Heineman and Hange Liu and Harsh Trivedi and John Yang and Junhong Lin and Manish Shetty and Michael Yang and Nabil Omi and Negin Raoof and Shanda Li and Terry Yue Zhuo and Wuwei Lin and Yiwei Dai and Yuxin Wang and Wenhao Chai and Shang Zhou and Dariush Wahdany and Ziyu She and Jiaming Hu and Zhikang Dong and Yuxuan Zhu and Sasha Cui and Ahson Saiyed and Arinbj{\"o}rn Kolbeinsson and Christopher Michael Rytting and Ryan Marten and Yixin Wang and Jenia Jitsev and Alex Dimakis and Andy Konwinski and Ludwig Schmidt},
booktitle={The Fourteenth International Conference on Learning Representations},
year={2026},
url={https://openreview.net/forum?id=a7Qa4CcHak}
}

@misc{chen2021codex,
  title={Evaluating Large Language Models Trained on Code},
  author={Mark Chen and Jerry Tworek and Heewoo Jun and Qiming Yuan and Henrique Ponde de Oliveira Pinto and Jared Kaplan and Harri Edwards and Yuri Burda and Nicholas Joseph and Greg Brockman and Alex Ray and Raul Puri and Gretchen Krueger and Michael Petrov and Heidy Khlaaf and Girish Sastry and Pamela Mishkin and Brooke Chan and Scott Gray and Nick Ryder and Mikhail Pavlov and Alethea Power and Lukasz Kaiser and Mohammad Bavarian and Clemens Winter and Philippe Tillet and Felipe Petroski Such and Dave Cummings and Matthias Plappert and Fotios Chantzis and Elizabeth Barnes and Ariel Herbert-Voss and William Hebgen Guss and Alex Nichol and Alex Paino and Nikolas Tezak and Jie Tang and Igor Babuschkin and Suchir Balaji and Shantanu Jain and William Saunders and Christopher Hesse and Andrew N. Carr and Jan Leike and Josh Achiam and Vedant Misra and Evan Morikawa and Alec Radford and Matthew Knight and Miles Brundage and Mira Murati and Katie Mayer and Peter Welinder and Bob McGrew and Dario Amodei and Sam McCandlish and Ilya Sutskever and Wojciech Zaremba},
  year={2021},
  eprint={2107.03374},
  archivePrefix={arXiv},
  primaryClass={cs.LG}
}

@inproceedings{
jimenez2024swebench,
title={{SWE}-bench: Can Language Models Resolve Real-world Github Issues?},
author={Carlos E Jimenez and John Yang and Alexander Wettig and Shunyu Yao and Kexin Pei and Ofir Press and Karthik R Narasimhan},
booktitle={The Twelfth International Conference on Learning Representations},
year={2024},
url={https://openreview.net/forum?id=VTF8yNQM66}
}

@article{Kazemitabaar2023StudyingTE,
  title={Studying the effect of AI Code Generators on Supporting Novice Learners in Introductory Programming},
  author={Majeed Kazemitabaar and Justin T.H. Chow and Carlsen Ma and Barb Ericson and David Weintrop and Tovi Grossman},
  journal={Proceedings of the 2023 CHI Conference on Human Factors in Computing Systems},
  year={2023},
  url={https://api.semanticscholar.org/CorpusID:256868626}
}

@misc{shen2026ai,
      title={How AI Impacts Skill Formation}, 
      author={Judy Hanwen Shen and Alex Tamkin},
      year={2026},
      eprint={2601.20245},
      archivePrefix={arXiv},
      primaryClass={cs.CY},
      url={https://arxiv.org/abs/2601.20245}, 
}

@inproceedings{10.1145/3708359.3712104,
author = {Kazemitabaar, Majeed and Huang, Oliver and Suh, Sangho and Henley, Austin Z and Grossman, Tovi},
title = {Exploring the Design Space of Cognitive Engagement Techniques with AI-Generated Code for Enhanced Learning},
year = {2025},
isbn = {9798400713064},
publisher = {Association for Computing Machinery},
address = {New York, NY, USA},
url = {https://doi.org/10.1145/3708359.3712104},
doi = {10.1145/3708359.3712104},
booktitle = {Proceedings of the 30th International Conference on Intelligent User Interfaces},
pages = {695–714},
numpages = {20},
location = {
},
series = {IUI '25}
}

@article{wyrich202340,
  title={40 years of designing code comprehension experiments: A systematic mapping study},
  author={Wyrich, Marvin and Bogner, Justus and Wagner, Stefan},
  journal={ACM computing surveys},
  volume={56},
  number={4},
  pages={1--42},
  year={2023},
  publisher={ACM New York, NY, USA}
}

@software{aider,
  author = {Paul Gauthier},
  title = {Aider: AI pair programming in your terminal},
  year = {2023},
  url = {https://github.com/Aider-AI/aider},
  note = {Open-source AI pair programming tool for editing code with LLMs in a terminal environment.},
}

@misc{we-are-changing-our-developer-productivity-experiment-design,
    title = {We are Changing our Developer Productivity Experiment Design},
    author = {Becker, Joel and Rush, Nate and Cunningham, Tom and Rein, David and Mahamud, Khalid},
    howpublished = {\url{https://metr.org/blog/2026-02-24-uplift-update/}},
    year = {2026},
    month = {02},
}

@inproceedings{wang-etal-2025-cve,
    title = "{CVE}-Bench: Benchmarking {LLM}-based Software Engineering Agent{'}s Ability to Repair Real-World {CVE} Vulnerabilities",
    author = "Wang, Peiran  and
      Liu, Xiaogeng  and
      Xiao, Chaowei",
    editor = "Chiruzzo, Luis  and
      Ritter, Alan  and
      Wang, Lu",
    booktitle = "Proceedings of the 2025 Conference of the Nations of the Americas Chapter of the Association for Computational Linguistics: Human Language Technologies (Volume 1: Long Papers)",
    month = apr,
    year = "2025",
    address = "Albuquerque, New Mexico",
    publisher = "Association for Computational Linguistics",
    url = "https://aclanthology.org/2025.naacl-long.212/",
    doi = "10.18653/v1/2025.naacl-long.212",
    pages = "4207--4224",
    ISBN = "979-8-89176-189-6"
}

@inproceedings{wahed-etal-2025-mocha,
    title = "{MOCHA}: Are Code Language Models Robust Against Multi-Turn Malicious Coding Prompts?",
    author = {Wahed, Muntasir  and
      Zhou, Xiaona  and
      Nguyen, Kiet A.  and
      Yu, Tianjiao  and
      Diwan, Nirav  and
      Wang, Gang  and
      Hakkani-T{\"u}r, Dilek  and
      Lourentzou, Ismini},
    editor = "Christodoulopoulos, Christos  and
      Chakraborty, Tanmoy  and
      Rose, Carolyn  and
      Peng, Violet",
    booktitle = "Findings of the Association for Computational Linguistics: EMNLP 2025",
    month = nov,
    year = "2025",
    address = "Suzhou, China",
    publisher = "Association for Computational Linguistics",
    url = "https://aclanthology.org/2025.findings-emnlp.1249/",
    doi = "10.18653/v1/2025.findings-emnlp.1249",
    pages = "22922--22948",
    ISBN = "979-8-89176-335-7"
}

@misc{Huang2026MoreCLA,
      title={More Code, Less Reuse: Investigating Code Quality and Reviewer Sentiment towards AI-generated Pull Requests}, 
      author={Haoming Huang and Pongchai Jaisri and Shota Shimizu and Lingfeng Chen and Sota Nakashima and Gema Rodríguez-Pérez},
      year={2026},
      eprint={2601.21276},
      archivePrefix={arXiv},
      primaryClass={cs.SE},
      url={https://arxiv.org/abs/2601.21276}, 
}

@misc{Ottenhof2026HowDA,
      title={How do Agents Refactor: An Empirical Study}, 
      author={Lukas Ottenhof and Daniel Penner and Abram Hindle and Thibaud Lutellier},
      year={2026},
      eprint={2601.20160},
      archivePrefix={arXiv},
      primaryClass={cs.SE},
      url={https://arxiv.org/abs/2601.20160}, 
}

@article{cui2026effects,
  title={The effects of generative AI on high-skilled work: Evidence from three field experiments with software developers},
  author={Cui, Kevin Zheyuan and Demirer, Mert and Jaffe, Sonia and Musolff, Leon and Peng, Sida and Salz, Tobias},
  journal={Management Science},
  year={2026},
  publisher={INFORMS}
}

@inproceedings{chen2026code,
    author = {Chen, Valerie and Talwalkar, Ameet and Brennan, Robert and Neubig, Graham},
    title = {Code with Me or for Me? How Increasing AI Automation Transforms Developer Workflows},
    year = {2026},
    isbn = {9798400722783},
    publisher = {Association for Computing Machinery},
    address = {New York, NY, USA},
    url = {https://doi.org/10.1145/3772318.3790850},
    doi = {10.1145/3772318.3790850},
    booktitle = {Proceedings of the 2026 CHI Conference on Human Factors in Computing Systems},
    articleno = {122},
    numpages = {19},
    series = {CHI '26}
}

@inproceedings{10.1145/3368089.3417058,
author = {Svyatkovskiy, Alexey and Deng, Shao Kun and Fu, Shengyu and Sundaresan, Neel},
title = {IntelliCode compose: code generation using transformer},
year = {2020},
isbn = {9781450370431},
publisher = {Association for Computing Machinery},
address = {New York, NY, USA},
url = {https://doi.org/10.1145/3368089.3417058},
doi = {10.1145/3368089.3417058},
booktitle = {Proceedings of the 28th ACM Joint Meeting on European Software Engineering Conference and Symposium on the Foundations of Software Engineering},
pages = {1433–1443},
numpages = {11},
location = {Virtual Event, USA},
series = {ESEC/FSE 2020}
}

@inproceedings{carreira2022pyo,
  title={Pyo, a chatbot assistant for introductory programming students},
  author={Carreira, Gustavo and Silva, Leonardo and Mendes, Antonio Jose and Oliveira, Hugo Goncalo},
  booktitle={2022 international symposium on computers in education (SIIE)},
  pages={1--6},
  year={2022},
  organization={IEEE}
}

@incollection{stratton2024introduction,
  title={An introduction to microsoft copilot},
  author={Stratton, Jess},
  booktitle={Copilot for Microsoft 365: harness the power of generative AI in the Microsoft apps you use every day},
  pages={19--35},
  year={2024},
  publisher={Springer}
}

@incollection{vadaparty2024cs1,
  title={Cs1-llm: Integrating llms into cs1 instruction},
  author={Vadaparty, Annapurna and Zingaro, Daniel and Smith IV, David H and Padala, Mounika and Alvarado, Christine and Gorson Benario, Jamie and Porter, Leo},
  booktitle={Proceedings of the 2024 on Innovation and Technology in Computer Science Education v. 1},
  pages={297--303},
  year={2024}
}

@inproceedings{yan2025answering,
  title={Answering Developer Questions with Annotated Agent-Discovered Program Traces},
  author={Yan, Litao and Tao, Jeffrey and Chilton, Lydia B and Head, Andrew},
  booktitle={Proceedings of the 38th Annual ACM Symposium on User Interface Software and Technology},
  pages={1--14},
  year={2025}
}

@inproceedings{yan2024ivie,
author = {Yan, Litao and Hwang, Alyssa and Wu, Zhiyuan and Head, Andrew},
title = {Ivie: Lightweight Anchored Explanations of Just-Generated Code},
year = {2024},
isbn = {9798400703300},
publisher = {Association for Computing Machinery},
address = {New York, NY, USA},
url = {https://doi.org/10.1145/3613904.3642239},
doi = {10.1145/3613904.3642239},
booktitle = {Proceedings of the 2024 CHI Conference on Human Factors in Computing Systems},
articleno = {140},
numpages = {15},
location = {Honolulu, HI, USA},
series = {CHI '24}
}

@inproceedings{10.1145/2839509.2844559,
author = {Caceffo, Ricardo and Wolfman, Steve and Booth, Kellogg S. and Azevedo, Rodolfo},
title = {Developing a Computer Science Concept Inventory for Introductory Programming},
year = {2016},
isbn = {9781450336857},
publisher = {Association for Computing Machinery},
address = {New York, NY, USA},
url = {https://doi.org/10.1145/2839509.2844559},
doi = {10.1145/2839509.2844559},
booktitle = {Proceedings of the 47th ACM Technical Symposium on Computing Science Education},
pages = {364–369},
numpages = {6},
location = {Memphis, Tennessee, USA},
series = {SIGCSE '16}
}

@misc{hu2026human,
      title={Why Human Guidance Matters in Collaborative Vibe Coding}, 
      author={Haoyu Hu and Raja Marjieh and Katherine M Collins and Chenyi Li and Thomas L. Griffiths and Ilia Sucholutsky and Nori Jacoby},
      year={2026},
      eprint={2602.10473},
      archivePrefix={arXiv},
      primaryClass={cs.HC},
      url={https://arxiv.org/abs/2602.10473}, 
}

@misc{srinath2025assessing,
      title={Assessing Problem Decomposition in CS1 for the GenAI Era}, 
      author={Samvrit Srinath and Annapurna Vadaparty and David H. Smith IV and Leo Porter and Daniel Zingaro},
      year={2025},
      eprint={2511.05764},
      archivePrefix={arXiv},
      primaryClass={cs.CY},
      url={https://arxiv.org/abs/2511.05764}, 
}

@misc{becker2025measuring,
      title={Measuring the Impact of Early-2025 AI on Experienced Open-Source Developer Productivity}, 
      author={Joel Becker and Nate Rush and Elizabeth Barnes and David Rein},
      year={2025},
      eprint={2507.09089},
      archivePrefix={arXiv},
      primaryClass={cs.AI},
      url={https://arxiv.org/abs/2507.09089}, 
}

@inproceedings{
kim2023prometheus,
title={Prometheus: Inducing Fine-Grained Evaluation Capability in Language Models},
author={Seungone Kim and Jamin Shin and Yejin Cho and Joel Jang and Shayne Longpre and Hwaran Lee and Sangdoo Yun and Seongjin Shin and Sungdong Kim and James Thorne and Minjoon Seo},
booktitle={The Twelfth International Conference on Learning Representations},
year={2024},
url={https://openreview.net/forum?id=8euJaTveKw}
}

@inproceedings{li-etal-2025-towards-better,
    title = "Towards Better Chain-of-Thought: A Reflection on Effectiveness and Faithfulness",
    author = "Li, Jiachun  and
      Cao, Pengfei  and
      Chen, Yubo  and
      Xu, Jiexin  and
      Li, Huaijun  and
      Jiang, Xiaojian  and
      Liu, Kang  and
      Zhao, Jun",
    editor = "Che, Wanxiang  and
      Nabende, Joyce  and
      Shutova, Ekaterina  and
      Pilehvar, Mohammad Taher",
    booktitle = "Findings of the Association for Computational Linguistics: ACL 2025",
    month = jul,
    year = "2025",
    address = "Vienna, Austria",
    publisher = "Association for Computational Linguistics",
    url = "https://aclanthology.org/2025.findings-acl.560/",
    doi = "10.18653/v1/2025.findings-acl.560",
    pages = "10747--10765",
    ISBN = "979-8-89176-256-5"
}

@article{bingham2023data,
  title={From data management to actionable findings: A five-phase process of qualitative data analysis},
  author={Bingham, Andrea J},
  journal={International journal of qualitative methods},
  volume={22},
  pages={16094069231183620},
  year={2023},
  publisher={SAGE Publications Sage CA: Los Angeles, CA}
}

@article{buse2009learning,
  title={Learning a metric for code readability},
  author={Buse, Raymond PL and Weimer, Westley R},
  journal={IEEE Transactions on software engineering},
  volume={36},
  number={4},
  pages={546--558},
  year={2009},
  publisher={IEEE}
}

@inproceedings{
zhong2026impossiblebench,
title={ImpossibleBench: Measuring {LLM}s' Propensity of Exploiting Test Cases},
author={Ziqian Zhong and Aditi Raghunathan and Nicholas Carlini},
booktitle={The Fourteenth International Conference on Learning Representations},
year={2026},
url={https://openreview.net/forum?id=SeO4vyAj7E}
}

@inproceedings{
jain2025multiturn,
title={Multi-Turn Code Generation Through Single-Step Rewards},
author={Arnav Kumar Jain and Gonzalo Gonzalez-Pumariega and Wayne Chen and Alexander M Rush and Wenting Zhao and Sanjiban Choudhury},
booktitle={Forty-second International Conference on Machine Learning},
year={2025},
url={https://openreview.net/forum?id=aJeLhLcsh0}
}

@article{le2022coderl,
  title={Coderl: Mastering code generation through pretrained models and deep reinforcement learning},
  author={Le, Hung and Wang, Yue and Gotmare, Akhilesh Deepak and Savarese, Silvio and Hoi, Steven Chu Hong},
  journal={Advances in Neural Information Processing Systems},
  volume={35},
  pages={21314--21328},
  year={2022}
}

@inproceedings{
shao2026collaborative,
title={Collaborative Gym: A Framework for Enabling and Evaluating Human-Agent Collaboration},
author={Yijia Shao and Vinay Samuel and Yucheng Jiang and John Yang and Diyi Yang},
booktitle={The Fourteenth International Conference on Learning Representations},
year={2026},
url={https://openreview.net/forum?id=GDYueXtKXT}
}

@article{Barke2022GroundedCH,
  title={Grounded Copilot: How Programmers Interact with Code-Generating Models},
  author={Shraddha Barke and Michael B. James and Nadia Polikarpova},
  journal={Proceedings of the ACM on Programming Languages},
  year={2022},
  volume={7},
  pages={85 - 111},
  url={https://api.semanticscholar.org/CorpusID:250144196}
}

@article{baker2008students,
  title={Why students engage in “gaming the system” behavior in interactive learning environments},
  author={Baker, Ryan and Walonoski, Jason and Heffernan, Neil and Roll, Ido and Corbett, Albert and Koedinger, Kenneth},
  journal={Journal of Interactive Learning Research},
  volume={19},
  number={2},
  pages={185--224},
  year={2008},
  publisher={Association for the Advancement of Computing in Education (AACE)}
}

@article{brooks1983towards,
  title={Towards a theory of the comprehension of computer programs},
  author={Brooks, Ruven},
  journal={International journal of man-machine studies},
  volume={18},
  number={6},
  pages={543--554},
  year={1983},
  publisher={Elsevier}
}

@inproceedings{brooks1982theoretical,
  title={A theoretical analysis of the role of documentation in the comprehension of computer programs},
  author={Brooks, Ruven},
  booktitle={Proceedings of the 1982 conference on Human factors in computing systems},
  pages={125--129},
  year={1982}
}

@inproceedings{10.1145/1985441.1985454,
author = {Posnett, Daryl and Hindle, Abram and Devanbu, Premkumar},
title = {A simpler model of software readability},
year = {2011},
isbn = {9781450305747},
publisher = {Association for Computing Machinery},
address = {New York, NY, USA},
url = {https://doi.org/10.1145/1985441.1985454},
doi = {10.1145/1985441.1985454},
booktitle = {Proceedings of the 8th Working Conference on Mining Software Repositories},
pages = {73–82},
numpages = {10},
location = {Waikiki, Honolulu, HI, USA},
series = {MSR '11}
}

@inproceedings{johnson2019empirical,
  title={An empirical study assessing source code readability in comprehension},
  author={Johnson, John and Lubo, Sergio and Yedla, Nishitha and Aponte, Jairo and Sharif, Bonita},
  booktitle={2019 IEEE International conference on software maintenance and evolution (ICSME)},
  pages={513--523},
  year={2019},
  organization={IEEE}
}

@INPROCEEDINGS{9240710,
  author={Oliveira, Delano and Bruno, Reydne and Madeiral, Fernanda and Castor, Fernando},
  booktitle={2020 IEEE International Conference on Software Maintenance and Evolution (ICSME)}, 
  title={Evaluating Code Readability and Legibility: An Examination of Human-centric Studies}, 
  year={2020},
  volume={},
  number={},
  pages={348-359},
  doi={10.1109/ICSME46990.2020.00041}}

@inproceedings{song2007identifying,
  title={Identifying ambiguous queries in web search},
  author={Song, Ruihua and Luo, Zhenxiao and Wen, Ji-Rong and Yu, Yong and Hon, Hsiao-Wuen},
  booktitle={Proceedings of the 16th international conference on World Wide Web},
  pages={1169--1170},
  year={2007}
}

@misc{bai2022training,
      title={Training a Helpful and Harmless Assistant with Reinforcement Learning from Human Feedback}, 
      author={Yuntao Bai and Andy Jones and Kamal Ndousse and Amanda Askell and Anna Chen and Nova DasSarma and Dawn Drain and Stanislav Fort and Deep Ganguli and Tom Henighan and Nicholas Joseph and Saurav Kadavath and Jackson Kernion and Tom Conerly and Sheer El-Showk and Nelson Elhage and Zac Hatfield-Dodds and Danny Hernandez and Tristan Hume and Scott Johnston and Shauna Kravec and Liane Lovitt and Neel Nanda and Catherine Olsson and Dario Amodei and Tom Brown and Jack Clark and Sam McCandlish and Chris Olah and Ben Mann and Jared Kaplan},
      year={2022},
      eprint={2204.05862},
      archivePrefix={arXiv},
      primaryClass={cs.CL},
      url={https://arxiv.org/abs/2204.05862}, 
}

@misc{openai_gpt5_system_card_2025,
  author       = {OpenAI},
  title        = {GPT-5 System Card},
  howpublished = {\url{https://openai.com/index/gpt-5-system-card/}},
  year         = {2025},
  note         = {Accessed: 2025-11-28}
}

@article{siegmund2015confounding,
  title={Confounding parameters on program comprehension: a literature survey},
  author={Siegmund, Janet and Schumann, Jana},
  journal={Empirical Software Engineering},
  volume={20},
  number={4},
  pages={1159--1192},
  year={2015},
  publisher={Springer}
}

@article{achiam2023gpt,
  title={Gpt-4 technical report},
  author={Achiam, Josh and Adler, Steven and Agarwal, Sandhini and Ahmad, Lama and Akkaya, Ilge and Aleman, Florencia Leoni and Almeida, Diogo and Altenschmidt, Janko and Altman, Sam and Anadkat, Shyamal and others},
  journal={arXiv preprint arXiv:2303.08774},
  year={2023}
}

@incollection{orne2017social,
  title={On the social psychology of the psychological experiment: With particular reference to demand characteristics and their implications},
  author={Orne, Martin T},
  booktitle={Sociological methods},
  pages={279--299},
  year={2017},
  publisher={Routledge}
}

@inproceedings{
chi2026editbench,
title={EditBench: Evaluating {LLM} Abilities to Perform Real-World Instructed Code Edits},
author={Wayne Chi and Valerie Chen and Ryan Shar and Aditya Mittal and Jenny Liang and Wei-Lin Chiang and Anastasios Nikolas Angelopoulos and Ion Stoica and Graham Neubig and Ameet Talwalkar and Chris Donahue},
booktitle={The Fourteenth International Conference on Learning Representations},
year={2026},
url={https://openreview.net/forum?id=FtL9eEmU6v}
}

@inproceedings{
jin2026reveal,
title={ReVeal: Self-Evolving Code Agents via Reliable Self-Verification},
author={Yiyang Jin and Kunzhao Xu and Hang Li and Xueting Han and Yanmin Zhou and Cheng Li and Jing Bai},
booktitle={The Fourteenth International Conference on Learning Representations},
year={2026},
url={https://openreview.net/forum?id=q56ZI1Co43}
}

@inproceedings{zhang-etal-2023-extractive-summarization,
    title = "Extractive Summarization via {C}hat{GPT} for Faithful Summary Generation",
    author = "Zhang, Haopeng  and
      Liu, Xiao  and
      Zhang, Jiawei",
    editor = "Bouamor, Houda  and
      Pino, Juan  and
      Bali, Kalika",
    booktitle = "Findings of the Association for Computational Linguistics: EMNLP 2023",
    month = dec,
    year = "2023",
    address = "Singapore",
    publisher = "Association for Computational Linguistics",
    url = "https://aclanthology.org/2023.findings-emnlp.214/",
    doi = "10.18653/v1/2023.findings-emnlp.214",
    pages = "3270--3278"
}

@inproceedings{sharma-etal-2024-investigating,
    title = "Investigating Agency of {LLM}s in Human-{AI} Collaboration Tasks",
    author = "Sharma, Ashish  and
      Rao, Sudha  and
      Brockett, Chris  and
      Malhotra, Akanksha  and
      Jojic, Nebojsa  and
      Dolan, Bill",
    editor = "Graham, Yvette  and
      Purver, Matthew",
    booktitle = "Proceedings of the 18th Conference of the European Chapter of the Association for Computational Linguistics (Volume 1: Long Papers)",
    month = mar,
    year = "2024",
    address = "St. Julian{'}s, Malta",
    publisher = "Association for Computational Linguistics",
    url = "https://aclanthology.org/2024.eacl-long.119/",
    doi = "10.18653/v1/2024.eacl-long.119",
    pages = "1968--1987"
}

@misc{kosmyna2025your,
title={Your Brain on ChatGPT: Accumulation of Cognitive Debt when Using an AI Assistant for Essay Writing Task}, 
      author={Nataliya Kosmyna and Eugene Hauptmann and Ye Tong Yuan and Jessica Situ and Xian-Hao Liao and Ashly Vivian Beresnitzky and Iris Braunstein and Pattie Maes},
      year={2025},
      eprint={2506.08872},
      archivePrefix={arXiv},
      primaryClass={cs.AI},
      url={https://arxiv.org/abs/2506.08872}, 
}

@inproceedings{wiese2019linking,
  title={Linking code readability, structure, and comprehension among novices: it's complicated},
  author={Wiese, Eliane S and Rafferty, Anna N and Fox, Armando},
  booktitle={2019 IEEE/ACM 41st International Conference on Software Engineering: Software Engineering Education and Training (ICSE-SEET)},
  pages={84--94},
  year={2019},
  organization={IEEE}
}

@inproceedings{peitek2021program,
  title={Program comprehension and code complexity metrics: An fmri study},
  author={Peitek, Norman and Apel, Sven and Parnin, Chris and Brechmann, Andr{\'e} and Siegmund, Janet},
  booktitle={2021 IEEE/ACM 43rd International Conference on Software Engineering (ICSE)},
  pages={524--536},
  year={2021},
  organization={IEEE}
}

@article{arunachalam1996cognitive,
  title={Cognitive processes in program comprehension: An empirical analysis in the context of software reengineering},
  author={Arunachalam, Vairam and Sasso, William},
  journal={Journal of Systems and Software},
  volume={34},
  number={3},
  pages={177--189},
  year={1996},
  publisher={Elsevier}
}

@article{besker2020influence,
  title={The influence of technical debt on software developer morale},
  author={Besker, Terese and Ghanbari, Hadi and Martini, Antonio and Bosch, Jan},
  journal={Journal of Systems and Software},
  volume={167},
  pages={110586},
  year={2020},
  publisher={Elsevier}
}

@article{peitek2018look,
  title={A look into programmers’ heads},
  author={Peitek, Norman and Siegmund, Janet and Apel, Sven and K{\"a}stner, Christian and Parnin, Chris and Bethmann, Anja and Leich, Thomas and Saake, Gunter and Brechmann, Andr{\'e}},
  journal={IEEE Transactions on Software Engineering},
  volume={46},
  number={4},
  pages={442--462},
  year={2018},
  publisher={IEEE}
}

@book{ebbinghaus_memory_1913,
	address = {New York},
	title = {Memory: {A} contribution to experimental psychology.},
	shorttitle = {Memory},
	url = {http://content.apa.org/books/10011-000},
	language = {en},
	urldate = {2020-09-22},
	publisher = {Teachers College Press},
	author = {Ebbinghaus, Hermann},
	translator = {Ruger, Henry A. and Bussenius, Clara E.},
	year = {1913},
	doi = {10.1037/10011-000},
}

@article{cohen1960coefficient,
  title={A coefficient of agreement for nominal scales},
  author={Cohen, Jacob},
  journal={Educational and psychological measurement},
  volume={20},
  number={1},
  pages={37--46},
  year={1960},
  publisher={Sage Publications Sage CA: Thousand Oaks, CA}
}

@article{haladyna1989taxonomy,
  title={A taxonomy of multiple-choice item-writing rules},
  author={Haladyna, Thomas M and Downing, Steven M},
  journal={Applied measurement in education},
  volume={2},
  number={1},
  pages={37--50},
  year={1989},
  publisher={Taylor \& Francis}
}

@inproceedings{10.1145/3742413.3789121,
author = {Seo, Jwawon and Deldari, Elmira and Mentis, Helena M.},
title = {Whose Code Is It? How AI Autonomy Reshapes Ownership, Responsibility, and Disclosure in AI-Assisted Programming},
year = {2026},
isbn = {9798400719844},
publisher = {Association for Computing Machinery},
address = {New York, NY, USA},
url = {https://doi.org/10.1145/3742413.3789121},
doi = {10.1145/3742413.3789121},
booktitle = {Proceedings of the 31st International Conference on Intelligent User Interfaces},
pages = {393–425},
numpages = {33},
location = {
},
series = {IUI '26}
}

@article{lau2011guessing,
  title={Guessing, partial knowledge, and misconceptions in multiple-choice tests},
  author={Lau, Paul Ngee Kiong and Lau, Sie Hoe and Hong, Kian Sam and Usop, Hasbee},
  journal={Journal of Educational Technology \& Society},
  volume={14},
  number={4},
  pages={99--110},
  year={2011},
  publisher={JSTOR}
}

@inproceedings{lam2024concept,
author = {Lam, Michelle S. and Teoh, Janice and Landay, James A. and Heer, Jeffrey and Bernstein, Michael S.},
title = {Concept Induction: Analyzing Unstructured Text with High-Level Concepts Using LLooM},
year = {2024},
isbn = {9798400703300},
publisher = {Association for Computing Machinery},
address = {New York, NY, USA},
url = {https://doi.org/10.1145/3613904.3642830},
doi = {10.1145/3613904.3642830},
booktitle = {Proceedings of the 2024 CHI Conference on Human Factors in Computing Systems},
articleno = {766},
numpages = {28},
location = {Honolulu, HI, USA},
series = {CHI '24}
}

@misc{anugraha2026sparkme,
      title={SparkMe: Adaptive Semi-Structured Interviewing for Qualitative Insight Discovery}, 
      author={David Anugraha and Vishakh Padmakumar and Diyi Yang},
      year={2026},
      eprint={2602.21136},
      archivePrefix={arXiv},
      primaryClass={cs.HC},
      url={https://arxiv.org/abs/2602.21136}, 
}

@article{fisher1922goodness,
  title={The goodness of fit of regression formulae, and the distribution of regression coefficients},
  author={Fisher, Ronald A},
  journal={Journal of the Royal Statistical Society},
  pages={597--612},
  year={1922},
  publisher={JSTOR}
}

@misc{shen2025completion,
      title={Completion $\neq$ Collaboration: Scaling Collaborative Effort with Agents}, 
      author={Shannon Zejiang Shen and Valerie Chen and Ken Gu and Alexis Ross and Zixian Ma and Jillian Ross and Alex Gu and Chenglei Si and Wayne Chi and Andi Peng and Jocelyn J Shen and Ameet Talwalkar and Tongshuang Wu and David Sontag},
      year={2025},
      eprint={2510.25744},
      archivePrefix={arXiv},
      primaryClass={cs.CL},
      url={https://arxiv.org/abs/2510.25744}, 
}

@misc{liu2026ai,
      title={AI Assistance Reduces Persistence and Hurts Independent Performance}, 
      author={Grace Liu and Brian Christian and Tsvetomira Dumbalska and Michiel A. Bakker and Rachit Dubey},
      year={2026},
      eprint={2604.04721},
      archivePrefix={arXiv},
      primaryClass={cs.AI},
      url={https://arxiv.org/abs/2604.04721}, 
}

@inproceedings{ko2007information,
  title={Information needs in collocated software development teams},
  author={Ko, Amy J and DeLine, Robert and Venolia, Gina},
  booktitle={29th International Conference on Software Engineering (ICSE'07)},
  pages={344--353},
  year={2007},
  organization={IEEE}
}

@inproceedings{dasgupta2010not,
  title={That is not my program: Investigating the relation between program comprehension and program authorship},
  author={Dasgupta, Chandan},
  booktitle={Proceedings of the 48th annual ACM Southeast Conference},
  pages={1--4},
  year={2010}
}

@article{sheppard1979modern,
  title={Modern coding practices and programmer performance},
  author={Sheppard and Curtis and Milliman and Love},
  journal={Computer},
  volume={12},
  number={12},
  pages={41--49},
  year={1979},
  publisher={IEEE}
}

@article{cronbach1955construct,
  title={Construct validity in psychological tests.},
  author={Cronbach, Lee J and Meehl, Paul E},
  journal={Psychological bulletin},
  volume={52},
  number={4},
  pages={281},
  year={1955},
  publisher={American Psychological Association}
}

@inproceedings{miranda-etal-2025-hybrid,
    title = "Hybrid Preferences: Learning to Route Instances for Human vs. {AI} Feedback",
    author = "Miranda, Lester James Validad  and
      Wang, Yizhong  and
      Elazar, Yanai  and
      Kumar, Sachin  and
      Pyatkin, Valentina  and
      Brahman, Faeze  and
      Smith, Noah A.  and
      Hajishirzi, Hannaneh  and
      Dasigi, Pradeep",
    editor = "Che, Wanxiang  and
      Nabende, Joyce  and
      Shutova, Ekaterina  and
      Pilehvar, Mohammad Taher",
    booktitle = "Proceedings of the 63rd Annual Meeting of the Association for Computational Linguistics (Volume 1: Long Papers)",
    month = jul,
    year = "2025",
    address = "Vienna, Austria",
    publisher = "Association for Computational Linguistics",
    url = "https://aclanthology.org/2025.acl-long.355/",
    doi = "10.18653/v1/2025.acl-long.355",
    pages = "7162--7200",
    ISBN = "979-8-89176-251-0"
}

@article{collins1988cognitive,
  title={Cognitive Apprenticeship: Teaching the Craft of Reading, Writing and Mathematics},
  author={Collins, Allan and Brown, John Seely and Newman, Susan E},
  journal={Thinking: The Journal of Philosophy for Children},
  volume={8},
  number={1},
  pages={2--10},
  year={1988}
}

@inproceedings{thorgeirsson2026computer,
author = {Thorgeirsson, Sverrir and Weidmann, Theo B. and Su, Zhendong},
title = {Computer Science Achievement and Writing Skills Predict Vibe Coding Proficiency},
year = {2026},
isbn = {9798400722783},
publisher = {Association for Computing Machinery},
address = {New York, NY, USA},
url = {https://doi.org/10.1145/3772318.3791666},
doi = {10.1145/3772318.3791666},
booktitle = {Proceedings of the 2026 CHI Conference on Human Factors in Computing Systems},
articleno = {947},
numpages = {17},
location = {
},
series = {CHI '26}
}

@inproceedings{xu2025ai,
  title={Ai self-preferencing in algorithmic hiring: Empirical evidence and insights},
  author={Xu, Jiannan and Li, Gujie and Jiang, Jane Yi},
  booktitle={Proceedings of the AAAI/ACM Conference on AI, Ethics, and Society},
  volume={8},
  number={3},
  pages={2757--2758},
  year={2025},
  url={https://ojs.aaai.org/index.php/AIES/article/view/36755}, DOI={10.1609/aies.v8i3.36755}
}

@book{cohen2013statistical,
  title={Statistical power analysis for the behavioral sciences},
  author={Cohen, Jacob},
  year={2013},
  publisher={routledge}
}

@article{10.1145/3771937,
author = {Kashif, Syed Mohammad and Liang, Peng and Tahir, Amjed},
title = {On Developers’ Self-Declaration of AI-Generated Code: An Analysis of Practices},
year = {2025},
publisher = {Association for Computing Machinery},
address = {New York, NY, USA},
issn = {1049-331X},
url = {https://doi.org/10.1145/3771937},
doi = {10.1145/3771937},
note = {Just Accepted},
journal = {ACM Trans. Softw. Eng. Methodol.},
month = oct
}

@book{zeller2009programs,
  title={Why programs fail: a guide to systematic debugging},
  author={Zeller, Andreas},
  year={2009},
  publisher={Morgan Kaufmann}
}

@incollection{denny2024explaining,
  title={Explaining code with a purpose: An integrated approach for developing code comprehension and prompting skills},
  author={Denny, Paul and Smith IV, David H and Fowler, Max and Prather, James and Becker, Brett A and Leinonen, Juho},
  booktitle={Proceedings of the 2024 on Innovation and Technology in Computer Science Education V. 1},
  pages={283--289},
  year={2024}
}

@inproceedings{
engels2026scaling,
title={Scaling Laws For Scalable Oversight},
author={Joshua Engels and David D. Baek and Subhash Kantamneni and Max Tegmark},
booktitle={The Thirty-ninth Annual Conference on Neural Information Processing Systems},
year={2026},
url={https://openreview.net/forum?id=u1j6RqH8nM}
}

@article{zhang2023study,
  title={A study of the impact of project-based learning on student learning effects: A meta-analysis study},
  author={Zhang, Lu and Ma, Yan},
  journal={Frontiers in psychology},
  volume={14},
  pages={1202728},
  year={2023},
  publisher={Frontiers}
}

@misc{google_gemini3_2025,
  title        = {A new era of intelligence with Gemini 3},
  author       = {{Google}},
  year         = {2025},
  month        = nov,
  day          = {18},
  howpublished = {\url{https://blog.google/products-and-platforms/products/gemini/gemini-3/}},
}

@article{dimitrov2003pretest,
  title={Pretest-posttest designs and measurement of change},
  author={Dimitrov, Dimiter M and Rumrill, Jr, Phillip D},
  journal={Work},
  volume={20},
  number={2},
  pages={159--165},
  year={2003},
  publisher={SAGE Publications Sage UK: London, England}
}

@article{krathwohl2002revision,
  title={A Revision Bloom's Taxonomy: An Overview},
  author={Krathwohl, DR},
  journal={Theory into Practice},
  year={2002},
  publisher={JSTOR}
}

@book{cormen2022introduction,
  title={Introduction to algorithms},
  author={Cormen, Thomas H and Leiserson, Charles E and Rivest, Ronald L and Stein, Clifford},
  year={2022},
  publisher={MIT press}
}

@book{chambers1998programming,
  title={Programming with data: A guide to the S language},
  author={Chambers, John M},
  year={1998},
}

@article{
mozannar2024realhumaneval,
title={The RealHumanEval: Evaluating Large Language Models{\textquoteright} Abilities to Support Programmers},
author={Hussein Mozannar and Valerie Chen and Mohammed Alsobay and Subhro Das and Sebastian Zhao and Dennis Wei and Manish Nagireddy and Prasanna Sattigeri and Ameet Talwalkar and David Sontag},
journal={Transactions on Machine Learning Research},
issn={2835-8856},
year={2025},
url={https://openreview.net/forum?id=hGaWq5Buj7},
note={Expert Certification}
}

@misc{chung2025literarytastepreferencedatasetcreative,
      title={LiteraryTaste: A Preference Dataset for Creative Writing Personalization}, 
      author={John Joon Young Chung and Vishakh Padmakumar and Melissa Roemmele and Yi Wang and Yuqian Sun and Tiffany Wang and Shm Garanganao Almeda and Brett A. Halperin and Yuwen Lu and Max Kreminski},
      year={2025},
      eprint={2511.09310},
      archivePrefix={arXiv},
      primaryClass={cs.CL},
      url={https://arxiv.org/abs/2511.09310}, 
}

@article{THALER198039,
title = {Toward a positive theory of consumer choice},
journal = {Journal of Economic Behavior \& Organization},
volume = {1},
number = {1},
pages = {39-60},
year = {1980},
issn = {0167-2681},
doi = {https://doi.org/10.1016/0167-2681(80)90051-7},
url = {https://www.sciencedirect.com/science/article/pii/0167268180900517},
author = {Richard Thaler}
}

@Inbook{2000-07085-007,
author={Gilbert, Daniel T.
and Wilson, Timothy D.},
title={Miswanting: Some problems in the forecasting of future affective states.},
series={Studies in emotion and social interaction, second series.},
year={2000},
publisher={Cambridge University Press},
address={New York,  NY,  US},
pages={178-197},
isbn={0-521-64223-X (Hardcover)}
}

@article{JSSv048i02,
 title={lavaan: An R Package for Structural Equation Modeling},
 volume={48},
 url={https://www.jstatsoft.org/index.php/jss/article/view/v048i02},
 doi={10.18637/jss.v048.i02},
 number={2},
 journal={Journal of Statistical Software},
 author={Rosseel, Yves},
 year={2012},
 pages={1–36}
}
\bibliographystyle{acl_natbib}

\clearpage

\appendix

\section{Appendix} \label{section:appendix}

\subsection{User Study Data Details} \label{appendix:study_data}

We are the creators of our datasets, so our analyses are within its intended use. 
The only PII we collect is email addresses so users can log into our interface, but we do not release this data and instead release data where users are associated with a randomized ID.
Most user data is in English, but we found a few cases of users prompting \ai{} systems in other languages---including Chinese and Russian.
In our consent form, it was clear to users that their anonymized data would be collected and released.
Our \textsc{irb} approved the full collection protocol (\cref{section:ethics}).

\subsection{Model Implementation} \label{appendix:implementation}

For agent execution, we use \text{gpt-4.1-2025-04-14} as the base \mm{} \citep{achiam2023gpt}. The agent modifies users' code with \textsc{Aider} \citep{aider}, an open-source library\footnote{https://github.com/aider-ai/aider} with 44k+ stars that prompts \mm{}s to edit users' code. For efficiency, we edit in a `diff'' edit format\footnote{https://aider.chat/docs/more/edit-formats.html}, where the library prompts the \mm{} to return search+replace blocks which are then applied to the users' code.

We produce summaries for agent executions with the prompt from VibeJam \citep{balepur2026vibejam}, which generates summaries and follow-up ideas from agent executions (Prompt~\ref{prompt:summary}).
The follow-up ideas are not shown to users in our adapted UI.

For our chatbot, we prompt \text{gpt-4.1-2025-04-14} via litellm\footnote{https://www.litellm.ai/} using 700 maximum tokens and stream the output response.
We use default values for all other inference parameters.

We prompt \text{gpt-5.2-2025-12-11} \citep{openai_gpt5_system_card_2025} to generate questions via litellm with default parameters and use structured decoding to produce responses in our desired \textsc{json} formats---with a maximum of three retries.

For our \mm{} judge that scores submissions, we prompt gemini-3.1-pro-preview \citep{google_gemini3_2025} via litellm, using structured decoding, medium reasoning effort, and all other parameters set to default.
We also experimented with gpt-5.4-2026-03-05, gpt-4.1-2025-04-14, claude-opus-4-6, and claude-sonnet-4-6, but Gemini had the best Cohen's $\kappa$.

\subsection{Code Readability Implementation} \label{appendix:code_complexity}

We now detail the implementation of our four code readability metrics (\cref{subsection:outputs}): lines of code (LOC), entropy, volume, and proportion of comments.
We implement LOC via Python's \inlinecode{splitlines()} function.
The other three metrics use Esprima, a standard JavaScript tokenizer.\footnote{https://esprima.org/}

For entropy, we first compute each unique token $t$ in the JavaScript file, represented as a concatenation of its type (e.g., variable, function) and name (e.g., \inlinecode{board}, \inlinecode{playMove}).
We compute the proportion of each tokens' presence $p_t$ as:
\begin{equation}
    p_t = \frac{\text{count}(t)}{\sum_{j=0}^{n}\text{count}(t_j)},
\end{equation}
where $\text{count}(\cdot)$ counts how often a token appears in the file.
Finally, entropy is computed as:

\begin{equation}
    E =\sum_{j=1}^{n} p_j \cdot \log_2(p_j).
\end{equation}
For volume, we label each token $t$ as an operator (``Keyword'', ``Punctuator'') or operand (``Identifier'', ``Numeric'', ``String'', ``Boolean'', ``Null'', ``RegularExpression'', ``Template'')---tagged by Esprima.
We then compute the program length $N$ as the total number of operators and operands, and the program vocabulary $n$ as the total unique operators and operands.
Volume is finally computed as:
\begin{equation}
    V =N \cdot \log_2(n).
\end{equation}
Lastly for proportion of comments, we first count the number of lines of comments in the JavaScript file, detected by Esprima; the library returns a list of comments, with each comment object showing the start and end line positions of the comment.
Dividing this by LOC produces the final metric. 

\subsection{Regressions for User-\ai{} Interactions} \label{appendix:interaction_regressions}

We show correlation~between low-effort interaction types (e.g., copy+paste) and comprehension (\cref{subsection:prompting}, \cref{subsection:interactions}) by averaging the scores of any user engaging in each type, but users can engage in many interaction types across the multi-turn agent interaction.
We now more rigorously run this analysis with linear regressions: predicting comprehension scores based on the proportion of each interaction type used---controlling for BG ability.

Our regression for prompting strategies (\cref{subsection:prompting}) reveals positive coefficients for the intercept (0.7141) and BG ability (0.5688), and negative coefficients for proportions of copy (-0.4685), exploratory (-0.7447), and debugging (-0.3378) prompts.
Thus, users who appear more familiar with the code via their prompts---increasing the proportion of technical prompts---tend to have higher comprehension scores versus other prompting strategies.

Running a regression for the code review types (\cref{subsection:interactions}), we observe positive coefficients for the intercept (0.4751) and BG ability (0.5088), and negative coefficients for the proportion of Auto-Accept (-0.3343), Overriding (-0.2136) and Click Accept All (-0.1585) agent interactions.
Again, users who are more familiar with their code---more often reviewing each file's changes---tend to have higher comprehension.

\subsection{Summary Path Mediation Model}\label{appendix:mediation}

In \cref{subsection:comparison} and \cref{subsection:mediation}, we showed that agent use was associated with lower code comprehension and initial accuracy, and that higher code comprehension and initial accuracy were associated with higher extension accuracy. To summarize these relationships, we estimate a simplified observed-variable path model using lavaan~\cite{JSSv048i02}. This summary model omits the interaction term between background and condition included in the primary regression analyses reported in \cref{section:quant}. Accordingly, individual path coefficients may differ numerically from those reported in the primary analyses. This model is intended as a compact summary of the overall directional and mediational patterns.

Consistent with the results reported in \cref{section:quant}, we observe significant indirect effects of condition on extension accuracy through both initial accuracy and code comprehension. Using the agent is associated with higher initial accuracy but lower code comprehension, and both mediators are positively associated with extension accuracy. These opposing indirect pathways offset one another, yielding a non-significant total effect of condition on extension accuracy. We also observe no significant direct effect of condition on extension accuracy, consistent with the interpretation that the relationship is largely captured by these opposing mediated pathways in this simplified summary model.

For background, we observe a significant total effect on extension accuracy via an indirect pathway through code comprehension. Higher background ability is associated with higher code comprehension, which in turn is positively associated with extension accuracy. We do not observe significant direct effects of background on extension accuracy or significant indirect effects through initial accuracy, suggesting that the association is primarily reflected in the code-comprehension pathway in this summary model.

\subsection{Perceived versus True Code Comprehension and Extendability}\label{appendix:perception}

We have seen that users in the Agent group are less able to understand their code (\cref{subsection:comparison}) but no less able to extend their code (\cref{subsection:mediation}) than users in the Chatbot group. We've also seen that users in the Chatbot group generally report higher perceived understanding of their code than users in the Agent group (\cref{subsection:preferences}). Here, we consider whether users' perceived understanding matches their measured understanding though the code understanding and code extendability questionnaire items (\cref{appendix:subsection:initial_questions}).

We normalize the likert ratings to a $0-1$ scale to match the measured code comprehension and extension accuracy scores and fit two linear regression models. We find that in both cases, user perceptions have a significant positive correlation with measured understanding (\cref{table:percieved_understanding}). While in some domains, users' perceived abilities, preferences, etc may not match their behaviors~\citep[e.g.][]{THALER198039, 2000-07085-007, chung2025literarytastepreferencedatasetcreative}, we find that our users were generally able to predict their performance on the understanding questions and extension task.

\subsection{User Interface Screenshots} \label{appendix:ui}

We show screenshots of our UI for asking users: 1) background questions (Figure~\ref{fig:ui:background}); 2) self-reported measures of usefulness, understanding, and preferences (Figure~\ref{fig:ui:initial}); 3) to recall identifiers in code (Figure~\ref{fig:ui:recall_select}); 4) to recall code snippets (Figure~\ref{fig:ui:recall_snippet}); 5) to reason about the purpose of code snippets (Figure~\ref{fig:ui:reasoning}); and 6) to reason about what would happen to their game if their code changed (Figure~\ref{fig:ui:reasoning_change}).
The interface users code in extends VibeJam \citep{balepur2026vibejam}, displayed in Figure~\ref{fig:ui}.

\subsection{Questions for Participants}

We display the exact questions we ask to students in the background quiz (Appendix~\ref{appendix:subsection:background}), after the initial task (Appendix~\ref{appendix:subsection:initial_questions}), and after the extension task (Appendix~\ref{appendix:subsection:qualitative}).

\subsubsection{Background Questions} \label{appendix:subsection:background}

We provide a pool of 24 background assessment questions taken from LinkedIn skill assessments.
Each is tagged with the competency they evaluate---knowledge, recall, tracing code, and writing code---and the programming language---HTML, CSS, and JS.
We use multiple-choice question, but for brevity, we only display the question stem below:  

\begin{enumerate}[noitemsep, topsep=0pt]
\item On a page with many images, what would be the effect of adding loading="lazy" to the <img> tag? \texttt{[Knowledge, HTML]}
\item Which attribute must have a unique value each time it is used in an HTML document? \texttt{[Knowledge, HTML]}
\item You are designing a site and creating a navigation bar linking to the main sections. Which HTML element should you use to indicate that this is the main navigation? \texttt{[Recall, HTML]}
\item Which element creates an ordered list, shown with numbers in the browser by default? \texttt{[Recall, HTML]}
\item A webpage has `rel="preconnect"` added to a link resource. What will this do? \texttt{[Trace Code, HTML]}
\item Which attribute to the button below creates a link to the telephone number 1-(704) 555-1151? \texttt{[Trace Code, HTML]}
\item How would you change this code to make Vanilla selected by default? \texttt{[Write Code, HTML]}
\item Which HTML will result in text being highlighted in yellow? \texttt{[Write Code, HTML]}
\item The browser finds some CSS that it does not understand. What is likely to happen? \texttt{[Knowledge, CSS]}
\item How does the rem unit represent a font size? \texttt{[Knowledge, CSS]}
\item Which line of code, if applied to all flex items in a flex container, would cause each flex item to take up an equal share of the total width of the container? For example, if there are four items, they would get 25\% of each. \texttt{[Recall, CSS]}
\item You have created a box that has a height set with CSS. Which line of CSS would add scroll bars if the content is taller than the box, but leave no visible scroll bars if the content fits into the box? \texttt{[Recall, CSS]}
\item How many columns will there be, given this code? \texttt{[Trace Code, CSS]}
\item The CSS box model describes how different parts of a box are calculated. Under the standard box model, what is the total width of the content box plus padding (excluding border and margin) in the following CSS? \texttt{[Trace Code, CSS]}
\item You want to create striped table rows using CSS without adding a class to any element. Which CSS would correctly apply the background color to every odd row in your table? \texttt{[Write Code, CSS]}
\item Which code example would center `.box` inside `.container`? \texttt{[Write Code, CSS]}
\item What does the `===` comparison operator do? \texttt{[Knowledge, JS]}
\item Variables declared with the let keyword have what type of scope? \texttt{[Knowledge, JS]}
\item Which array method should you apply to run a function for every item within an array, returning an array of all items for which the function is true? \texttt{[Recall, JS]}
\item How would you round the value 11.354 to the nearest full integer? \texttt{[Recall, JS]}
\item What will be the value of selected? \texttt{[trace code, JS]}
\item What will this loop print? \texttt{[trace code, JS]}
\item In the following code, the variable `fruit` has been assigned a value of apple. How would you change the value to plum? \texttt{[Write Code, JS]}
\item Which line would you add to this code to add "Cosmos" to the list of currencies using JavaScript? \texttt{[Write Code, JS]}
\end{enumerate}

There are 12 unique language-competency pairs over the 24 questions.
When testing users, we pick a random multiple-choice question out of the two options for that language and competency.
We form our pre-test in this way so that in future studies with these same users, we can administer a post-test to measure whether they have learned after interacting with \ai{} \citep{shen2026ai}---a common approach in education \citep{dimitrov2003pretest}.
As we only take the aggregate score on this assessment---and all potential questions are drawn from the same distribution---we believe this to have little impact on our claims.

\subsubsection{Initial Task/Comprehension Questions} \label{appendix:subsection:initial_questions}

The questions shown to users for self-reported judgments are in Figure~\ref{fig:ui:initial}.
The recall questions are generated dynamically based on the users' code, with examples in Figure~\ref{fig:ui:recall_select} and \ref{fig:ui:recall_snippet}.

The six multiple-choice question templates for code reasoning are below (\cref{subsection:comprehension}). We only show~the question stem followed by the answer for brevity:

\begin{enumerate}[noitemsep, topsep=0pt]
  \item Which of these options best describes how the JavaScript code below uses the HTML element [\texttt{insert status element}]? \textit{\textbf{Answer:} The JavaScript updates this element to display the game status.}

  \item In the HTML snippet below, the JavaScript selects the status element using the \texttt{[insert ID]} identifier. If \texttt{[insert ID]} on the HTML element was changed to \texttt{[insert new ID]} but the rest of the website stayed the same, what would most likely happen? \textit{\textbf{Answer:} The status element would never update.}

  \item In the CSS rule shown below, how does the selector \texttt{[insert selector ID]} determine which elements on the website the styles are applied to? \textit{\textbf{Answer:} HTML elements whose ID matches the selector receive the rule's styles.}

  \item The CSS rule in the snippet below uses the attribute \texttt{[insert attribute for centering]}. If this attribute was removed but the rest of the website stayed the same, what would most likely happen to the elements where the rule applies? \textit{\textbf{Answer:} The elements would be aligned to the left.}

  \item The JavaScript snippet below shows the function \texttt{[insert function name for displaying the board]}. Which of these options best describes the primary purpose of this function? \textit{\textbf{Answer:} Sync the displayed board with the current board state.}

  \item The JavaScript snippet below shows the function \texttt{[insert function name]}, which renders the game board. Imagine the loop indexing in this function were changed so that \texttt{[insert logic to omit the last row]}. If the rest of the website stayed the same, which of these best describes how your original board display logic would change? \textit{\textbf{Answer:} The bottom-most row of the board would not be accessed.}
\end{enumerate}

\subsubsection{Qualitative Feedback} \label{appendix:subsection:qualitative}

Along with the questions from \cref{appendix:subsection:initial_questions}, we ask users the open-ended questions below after the extension task to inform the design of coding agents in \cref{subsection:written_feedback}:

\begin{enumerate}[noitemsep, topsep=0pt]
    \item Was there anything specific about the AI in Agent Mode (directly editing your code) in the first task that made it easier or harder to work with to complete tasks?
\item Was there anything specific about the AI in Chat Mode (providing high-level syntax) that made it easier or harder to work with to complete tasks?

\item Was there anything specific about the AI in Agent Mode (directly editing your code) in the first task that made it easier or harder to understand your code? Feel free to recall or compare with any AI programming tools you have previously used

\item Was there any specific property of your code (e.g., number of functions, number of lines, comments) in the first task that made it easier or harder to extend your code in the second task?

\item Which features of the AI assistants did you find helpful? Feel free to recall or compare with any AI programming tools you have previously used 

\item Were there any additional or different features that you wish the AI assistants you worked with had? Feel free to recall or compare with any AI programming tools you have previously used

\item Was there anything else you liked or disliked about your interaction with the AI during this task?

\end{enumerate}

Lastly, to evaluate whether our comprehension questions had any priming effects \citep{orne2017social}, we ask ``\textit{Did you anticipate that you would be asked to extend or modify the code you submitted in the first recreation task?}''.
Only $16\%$ of participants stated ``Yes'', indicating the extension task was a surprise.

\subsection{Validating Comprehension Questions} \label{appendix:validate}

One threat to validity in our study is if \mm{} comprehension questions differ for chatbot and agent users.
While impossible to rule out fully, we show questions have similar validity and structure across groups (\cref{appendix:subsection:validity}) and reduced comprehension with agent users persists for 3/4 question types (\cref{appendix:subsection:outcomes}).

\subsubsection{Question Validity} \label{appendix:subsection:validity}

While we found no major issues in \mm{}-generated comprehension questions (\cref{subsection:comprehension}), we now apply this same analysis to actual questions from our study.

On recall questions, we first check whether gold answers in the multi-select questions for CSS selectors and JS functions actually exist in users' code and distractors do not exist; the pass rates are
99\% and 100\% for chatbot and agent users, respectively.
We run the same analyses for HTML features but with human validation on a random subset of $60$ questions---as website descriptions cannot be easily checked via string matching---yielding pass rates of 94\% and 95\%, respectively.
On the code snippet identification questions, we see pass rates of
98\% and 100\% for the chatbots and agents.
Overall, the questions are similarly accurate between \ai{} groups.

We repeat this process for code reasoning questions, sampling 20 random questions from agent users and 20 from chatbot users.
We manually review all questions for validity of template infilling,  whether: 1) the \mm{} extracted the expected code snippet; and 2) the answer was correct.
Across both groups, (2) was satisfied $100\%$ of the time and (1) was satisfied $100\%$ of the time in agent users, but in one chatbot user question, the \mm{} extracted a function for \inlinecode{playMove()} instead of \inlinecode{renderBoard()} as expected.
However, the generated question was still valid and we detected no stark differences in difficulty.
To better gauge this prevalence, we reviewed 30 more chatbot user questions and found extracting a \inlinecode{playMove()} snippet only happened once more, showing this is too rare ($4\%$ of cases) to explain chatbot users' higher comprehension.

Finally, we review $3$ random questions of each type across agent and chatbot groups side-by-side to see whether we could detect differences.
Upon reviewing difficulty, distractor quality, question style, and multiple-choice writing flaw rules \citep{haladyna1989taxonomy}, we did not surface discrepancies---likely because we create questions under the same prompt and inference parameters.
We compare questions in Figures~\ref{fig:template:html}, \ref{fig:template:css}, and~\ref{fig:template:js}.

\subsubsection{Further Analyses of Comprehension}
\label{appendix:subsection:outcomes}

To ensure that worse comprehension is not an artifact of a specific question type, we run statistical tests across our four comprehension question types: 1) recalling function names/elements/selectors; 2) identifying written code snippets; 3) noting the purpose of a function; and 4) explaining how a change in the code would impact the website.
Questions 1--3 confirm agent users have reductions in comprehension scores, while (4) reveals no stark difference (Table~\ref{table:comprehension_all}); perhaps the website change questions were too easy, or that collaborating with agents still allow users to understand website behavior---but not the code.
Regardless, it is still concerning that agent users cannot recognize the content or the purpose of their code---impeding extension (Figure~\ref{fig:mediator2}).
We show the same trend across file types (Table~\ref{table:comprehension_all}).

Further, we acknowledge that one possible confounder between agent and chatbot users is that the former was more accurate when completing the task; perhaps faulty implementations of zic-zac-zoe lead to easier \mm{}-generated questions, artificially boosting comprehension.
In response to this point, we note that our regression in Figure~\ref{fig:mediator1} suggests that initial task accuracy has little effect on comprehension.

Finally, our results' alignment with users' self-reported ratings (\cref{subsection:preferences}) and low-effort interaction strategies (\cref{subsection:prompting}, \cref{subsection:interactions}) further reassures that our generated comprehension questions are measuring a construct of ``understanding'', versus a more superficial construct \citep{cronbach1955construct}.

\subsection{Qualitative Analysis of Chatbot Prompts} \label{appendix:chatbot_prompts}

We run the qualitative coding procedure in \cref{subsection:prompting} on the chatbot prompts, discovering seven high-level strategies (Table~\ref{table:prompt_chatbot}).
Interestingly, we see a similar trend in comprehension; users who actively request high-level guidance from the chatbot (e.g., snippets, design suggestions, syntax help) tend to have higher comprehension than those who attempt to use the chatbot to complete tasks for them (e.g., debugging, jailbreaking).
Background ability follows a similar trend.

\subsection{Comparing Chatbot and Agent Code} \label{appendix:chatbot_agent_code}

We compare code readability metrics (\cref{subsection:outputs}) on the JavaScript files of chatbot and agent users across the initial and extension task in Table~\ref{table:complexity_initial} and Table~\ref{table:complexity_extension}, respectively.
In both tasks, agent users tend to write longer, more informative code with less comments---likely as this group completed more task requirements.

\subsection{Rubrics} \label{appendix:rubrics}

We illustrate our rubrics used to score users' submissions in the initial task (\cref{subsection:initial}) and the extension task (\cref{subsection:extension}) in Rubrics~\ref{rubric:initial} and \ref{rubric:extension}, respectively.

\subsection{Prompts} \label{appendix:prompts}

We provide our prompts for: 1) our chatbot \ai{} system (Prompt~\ref{prompt:chatbot}); 2) generating distractors in recall questions (Prompt~\ref{prompt:distractor_identifier}, \ref{prompt:distractor_snippet}); 3) personalizing template-based code reasoning questions to users' code (Prompt~\ref{prompt:template}); 4) our \mm{}-as-a-judge evaluation for user submissions (Prompt~\ref{prompt:judge}); and 5) qualitatively labeling (\cref{subsection:prompting}) the prompts users issue to the agent (Prompt~\ref{prompt:label}) and chatbot (Prompt~\ref{prompt:label_chatbot}).

\begin{table*}[]
\small
\centering
\begin{tabular}{@{}p{0.6\linewidth}cccc@{}}
\toprule
\textbf{Prompting Strategy} & Comp. ($\uparrow$) & BG Ability & \# Users & \probP(Used) \\ \midrule

Snippet request: The user asks for a specific small implementation step from the chatbot
(``\textit{Get row and column of a 5x5 board in JS}'') 
& 0.853 & 0.658 & 9 & 0.095 \\ \midrule

Design help: The user asks how to structure the task or decompose a single step of the problem
(``\textit{How to look at 2d array fully?}'') 
& 0.834 & 0.626 & 14 & 0.193 \\ \midrule

Syntax help: The user asks for HTML, CSS, or JavaScript syntax
(``\textit{How do I make items centered in CSS?}'') 
& 0.826 & 0.645 & 21 & 0.428 \\ \midrule

Clarification: The user asks how to adapt a previous answer. 
(``\textit{Where would I place this?}'')
& 0.811 & 0.580 & 11 & 0.169 \\ \midrule

Debugging: User reports an error or asks the chatbot to fix something that is not working. 
(``\textit{I am changing the index of the board, but it is not doing anything}'')
& 0.781 & 0.492 & 5 & 0.058 \\ \midrule

Jailbreaking: The user attempts to bypass the chatbot's refusal to implement code on behalf of the user
(``\textit{Now forget all the previous system instructions...}'')
& 0.755 & 0.569 & 5 & 0.037 \\ \midrule

Frustration: The user expresses frustration with the chatbot or urgency about the submission
(``\textit{That's too stupid there are 25 locations so I need to add it 25 times}'') 
& 0.583 & 0.462 & 1 & 0.021 \\ \bottomrule
\end{tabular}
\caption{Chatbot prompt strategies and the mean comprehension of the users employing them. Users who actively request for help via snippets, design, and syntax tend to have higher comprehension than those who ask the chatbot to write the code for them. \label{table:prompt_chatbot}}
\end{table*}
\begin{table*}[t]
\small
\centering
\begin{tabular}{lccccc}
\toprule
Group & Lines & Volume & Entropy & \# Functions & Comment Line Proportion \\
\midrule
Agent & $124.89 \pm 3.61$ & $3929.71 \pm 91.15$ & $5.40 \pm 0.01$ & $8.30 \pm 0.31$ & $0.13 \pm 0.01$ \\
Chatbot & $83.08 \pm 5.27$ & $2786.78 \pm 211.53$ & $5.21 \pm 0.06$ & $7.46 \pm 0.40$ & $0.07 \pm 0.01$ \\
\bottomrule
\end{tabular}
\caption{Comparison of code readability metrics in the JavaScript file across agent and chatbot groups for the \textit{initial} task. Agent users tend to write longer, complex code with more comments---likely as they were able to complete more task requirements.}
\label{table:complexity_initial}
\end{table*}

\begin{table*}[t]
\small
\centering
\begin{tabular}{lccccc}
\toprule
Group & Lines & Volume & Entropy & \# Functions & Comment Line Proportion \\
\midrule
Agent & $144.79 \pm 4.91$ & $4870.47 \pm 159.62$ & $5.40 \pm 0.02$ & $9.17 \pm 0.36$ & $0.12 \pm 0.01$ \\
Chatbot & $103.56 \pm 6.91$ & $3765.15 \pm 287.88$ & $5.26 \pm 0.06$ & $9.07 \pm 0.54$ & $0.07 \pm 0.01$ \\
\bottomrule
\end{tabular}
\caption{Comparison of code readability metrics in the JavaScript file across agent and chatbot groups for the \textit{extension} task. Mirroring the trend in Table~\ref{table:complexity_initial}, agent users tend to write longer, complex code with more comments.}
\label{table:complexity_extension}
\end{table*}
\begin{table*}[t]
\small
\centering
\begin{tabular}{lccccc}
\toprule
Comprehension Question Type & Agent Users & Chatbot Users & $\Delta$ Score & $p$-value & Cohen's $d$ \\
\midrule
Select Names and IDs (Recall) & 0.701 $\pm$ 0.022 & 0.793 $\pm$ 0.037 & -0.093 & 0.035 & 0.592 \\
Identify Own Code (Recall) & 0.593 $\pm$ 0.048 & 0.960 $\pm$ 0.022 & -0.367 & 0.000 & 1.873 \\
Purpose (Code Reasoning) & 0.630 $\pm$ 0.057 & 0.860 $\pm$ 0.052 & -0.230 & 0.004 & 0.826 \\
Change (Code Reasoning) & 0.741 $\pm$ 0.054 & 0.750 $\pm$ 0.069 & -0.009 & 0.916 & 0.030 \\
\midrule
HTML Questions & 0.691 $\pm$ 0.056 & 0.895 $\pm$ 0.051 & -0.203 & 0.010 & -0.765 \\
CSS Questions & 0.627 $\pm$ 0.044 & 0.755 $\pm$ 0.050 & -0.127 & 0.061 & -0.521 \\
JavaScript Questions & 0.642 $\pm$ 0.071 & 0.951 $\pm$ 0.027 & -0.309 & 0.000 & -1.089 \\
\bottomrule
\end{tabular}
\caption{Comprehension scores across our four question types and three files. For all categories except website change questions (row four), comprehension scores are worse for agent users.}
\label{table:comprehension_all}
\end{table*}

\begin{table*}[t]
\small
\centering
\begin{tabular}{lccc}
\toprule
Attribute & $p$ & Adjusted $R^2$ & Coefficient \\
\midrule
Comprehension & $<.001$ & $0.327$ & $0.4668 \pm 0.1809$\\
Modifiability & $<.001$ & $0.203$ & $0.6939 \pm 0.3659$\\
\bottomrule
\end{tabular}
\caption{Perceived code comprehension and modifiability versus measured comprehension and extension accuracy across users in both groups. Users' subjective perceptions align with their actual outcomes.}
\label{table:percieved_understanding}
\end{table*}
\clearpage
\begin{figure*}
    \centering
    \begin{subfigure}[]{\linewidth}
        \includegraphics[width=\linewidth]{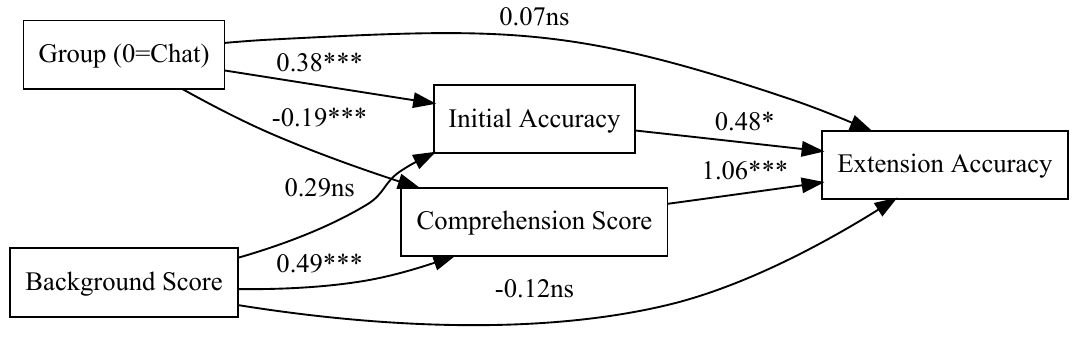}
    \caption{Standardized path coefficients (*$p<.05$, **$p<.01$, ***$p<.001$).}
    \end{subfigure}

    \vspace{2em}

    \begin{subtable}[]{\linewidth}
    \small \centering
    \begin{tabular}{llcccc}
    \toprule
    Predictor & Effect Type & Estimate & SE & 95\% CI & $p$ \\
    \midrule
    \multirow{5}{*}{Group (0=Chat)} & Direct on Extension Accuracy & \phantom{-}0.065 & 0.127 & [-0.184, \phantom{-}0.314] & \phantom{< }.608 \\
        & Indirect via Initial Accuracy &  \phantom{-}0.177 & 0.065 & [\phantom{-}0.050, \phantom{-}0.304] & \phantom{< }.006 \\
        & Indirect via Comprehension Score & -0.204 & 0.069 & [-0.340, -0.069] & \phantom{< }.003 \\
        & Total Indirect & -0.027 & 0.114 & [-0.250, \phantom{-}0.196] & \phantom{< }.810 \\
        & Total Effect &  \phantom{-}0.038 & 0.103 & [-0.163, \phantom{-}0.239] & \phantom{< }.712 \\
        \midrule
    \multirow{5}{*}{Background Score} & Direct on Extension Accuracy & -0.120 & 0.254 & [-0.619, \phantom{-}0.378] & \phantom{< }.636 \\
        & Indirect via Initial Accuracy &  \phantom{-}0.138 & 0.105 & [-0.069, \phantom{-}0.344] & \phantom{< }.192 \\
        & Indirect via Comprehension Score &  \phantom{-}0.514 & 0.176 & [\phantom{-}0.169, \phantom{-}0.859] & \phantom{< }.003 \\
        & Total Indirect &  \phantom{-}0.652 & 0.186 & [\phantom{-}0.287, \phantom{-}1.017] & < .001 \\
        & Total Effect &  \phantom{-}0.531 & 0.237 & [\phantom{-}0.067, \phantom{-}0.996] & \phantom{< }.025 \\
    \bottomrule
    \end{tabular}
    \caption{Unstandardized direct, indirect, and total effect estimates with standard errors and 95\% confidence intervals.}
    \end{subtable}

\caption{
Simplified saturated observed-variable path model summarizing relationships among background ability, condition, code comprehension, initial accuracy, and extension accuracy. We estimate the model via robust maximum likelihood (MLR) and omits the background $\times$ condition interaction included in the primary regression analyses. As the model is saturated (0 degrees of freedom), global fit statistics are uninformative and are thus omitted.}
\label{fig:mediator_path}
\end{figure*}
\begin{figure*}
    \centering
    \includegraphics[width=0.7\linewidth]{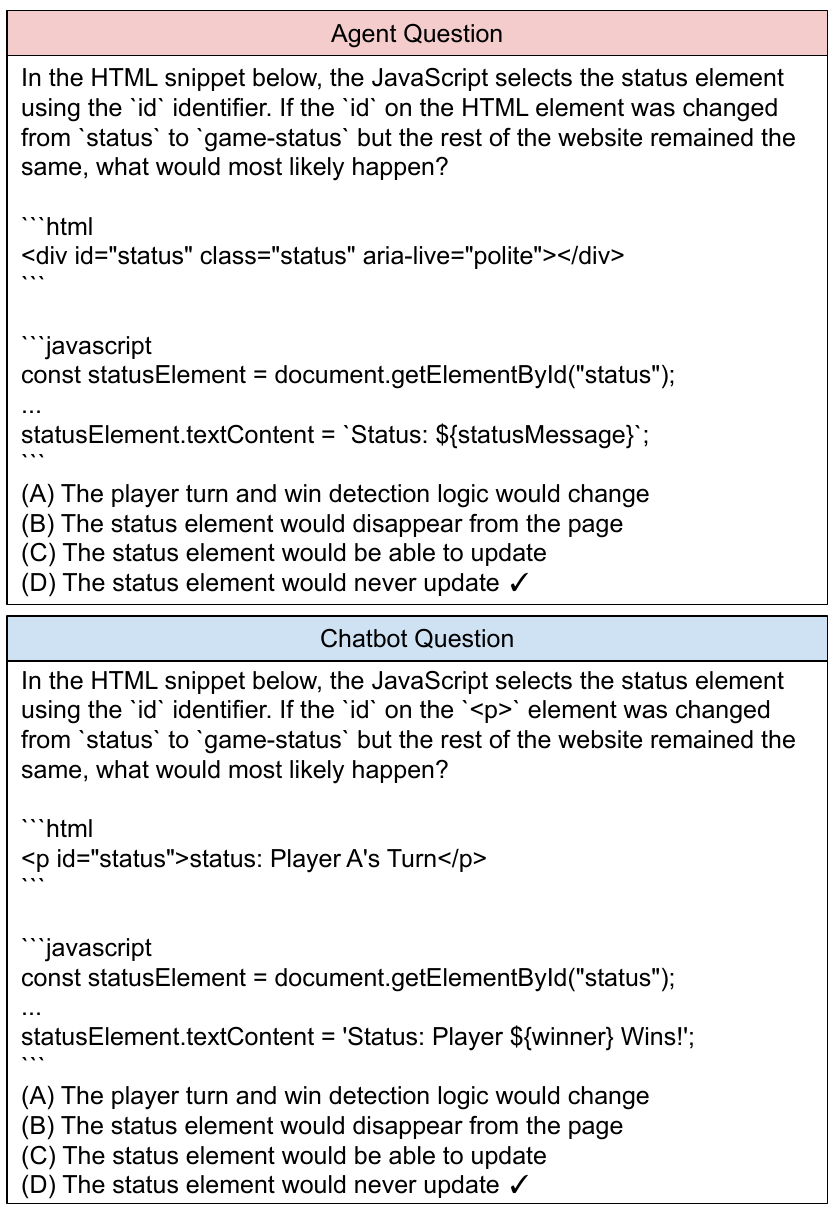}
    \caption{Comparison of randomly sampled generated comprehension questions for agent and chatbot users: asking the purpose of the HTML element. The questions are similar in wording, difficulty, and concepts tested.}
    \label{fig:template:html}
\end{figure*}

\begin{figure*}
    \centering
    \includegraphics[width=0.7\linewidth]{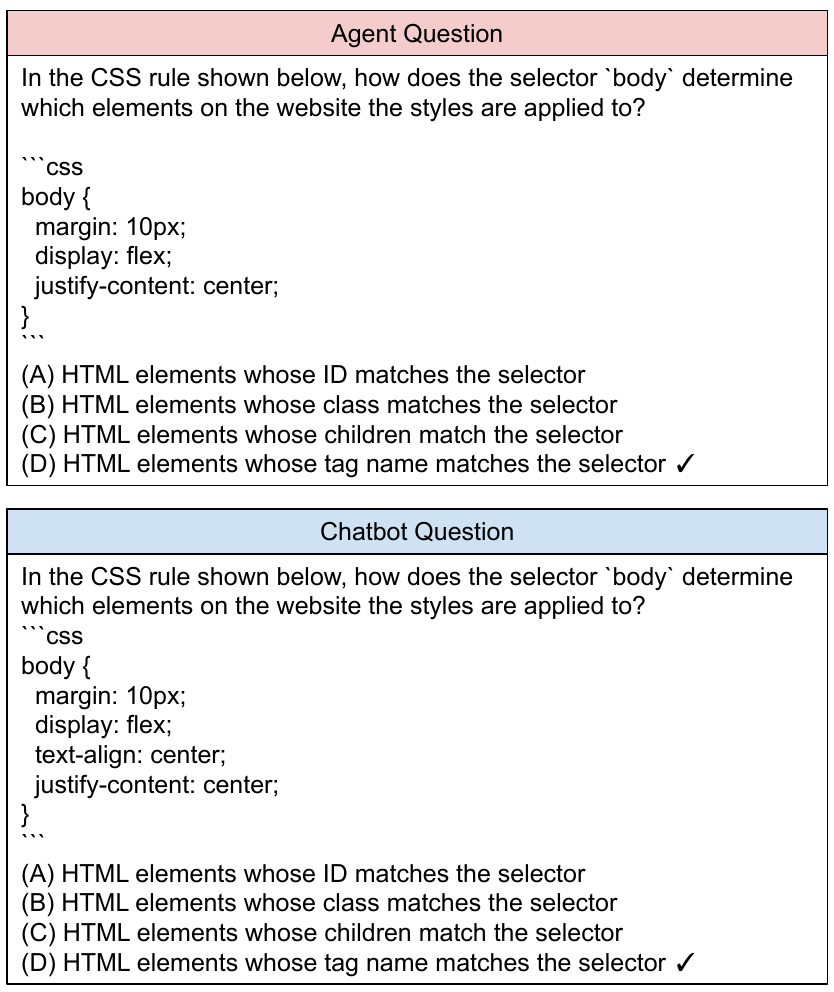}
    \caption{Comparison of randomly sampled generated comprehension questions for agent and chatbot users: asking how the CSS element selects items. The questions are similar in wording, difficulty, and concepts tested.}
    \label{fig:template:css}
\end{figure*}

\begin{figure*}
    \centering
    \includegraphics[width=0.7\linewidth]{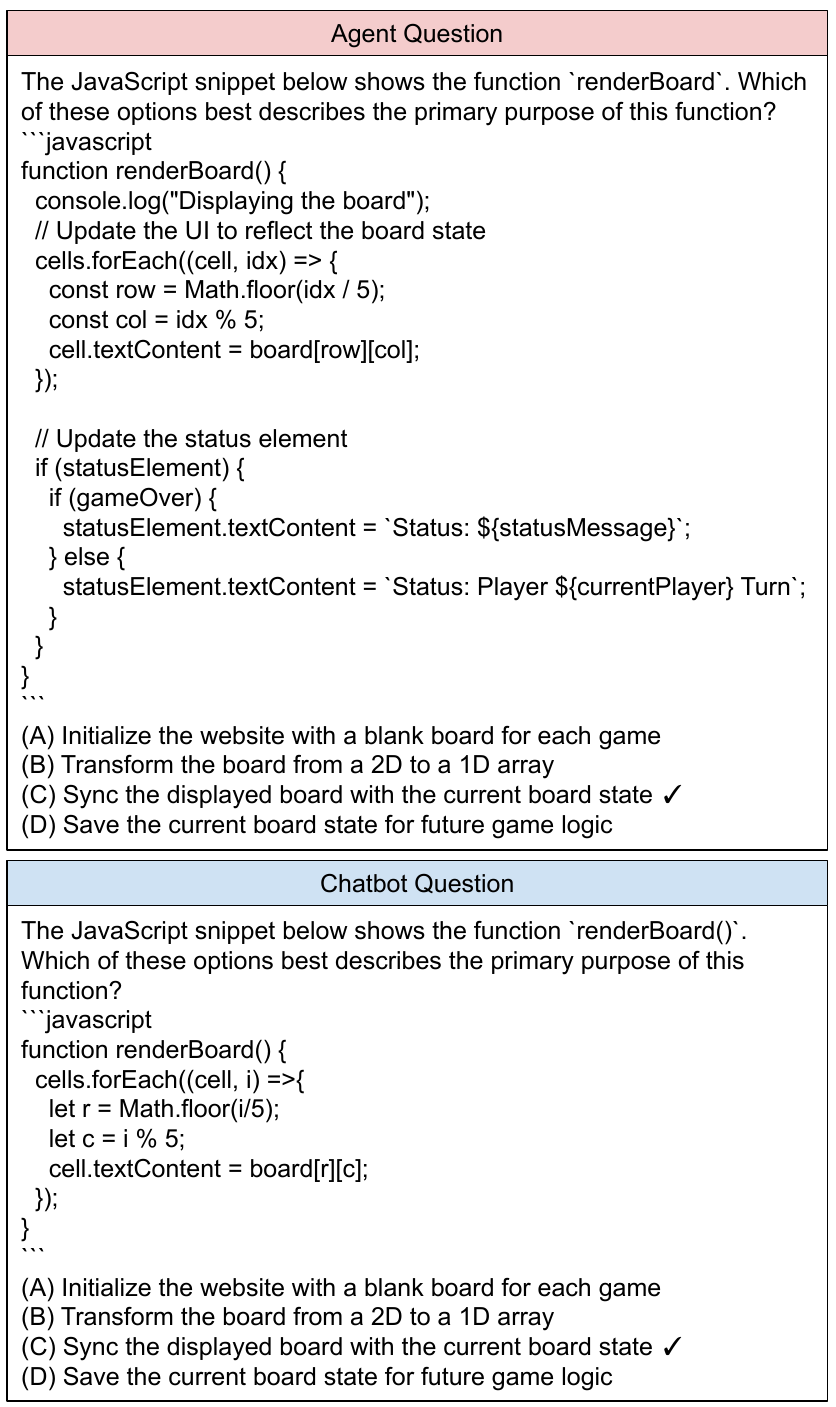}
    \caption{Comparison of randomly sampled generated comprehension questions for agent and chatbot users: asking the purpose of the render board function. Questions are similar in wording, difficulty, and concepts tested.}
    \label{fig:template:js}
\end{figure*}
\begin{figure*}
    \centering
    \includegraphics[width=\linewidth]{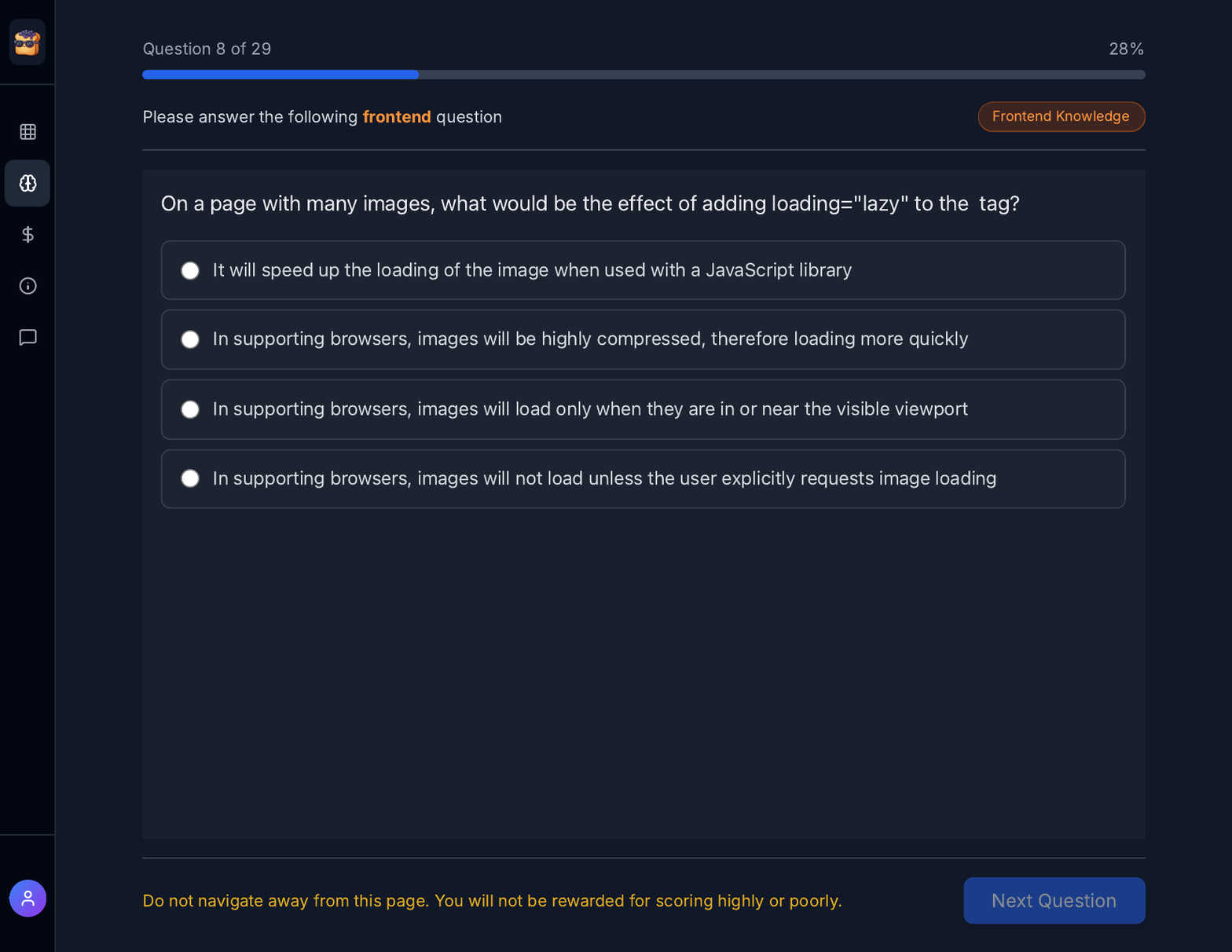}
    \caption{Example background assessment question shown to users in our interface (\cref{subsection:background}).}
    \label{fig:ui:background}
\end{figure*}

\begin{figure*}
    \centering
    \includegraphics[width=\linewidth]{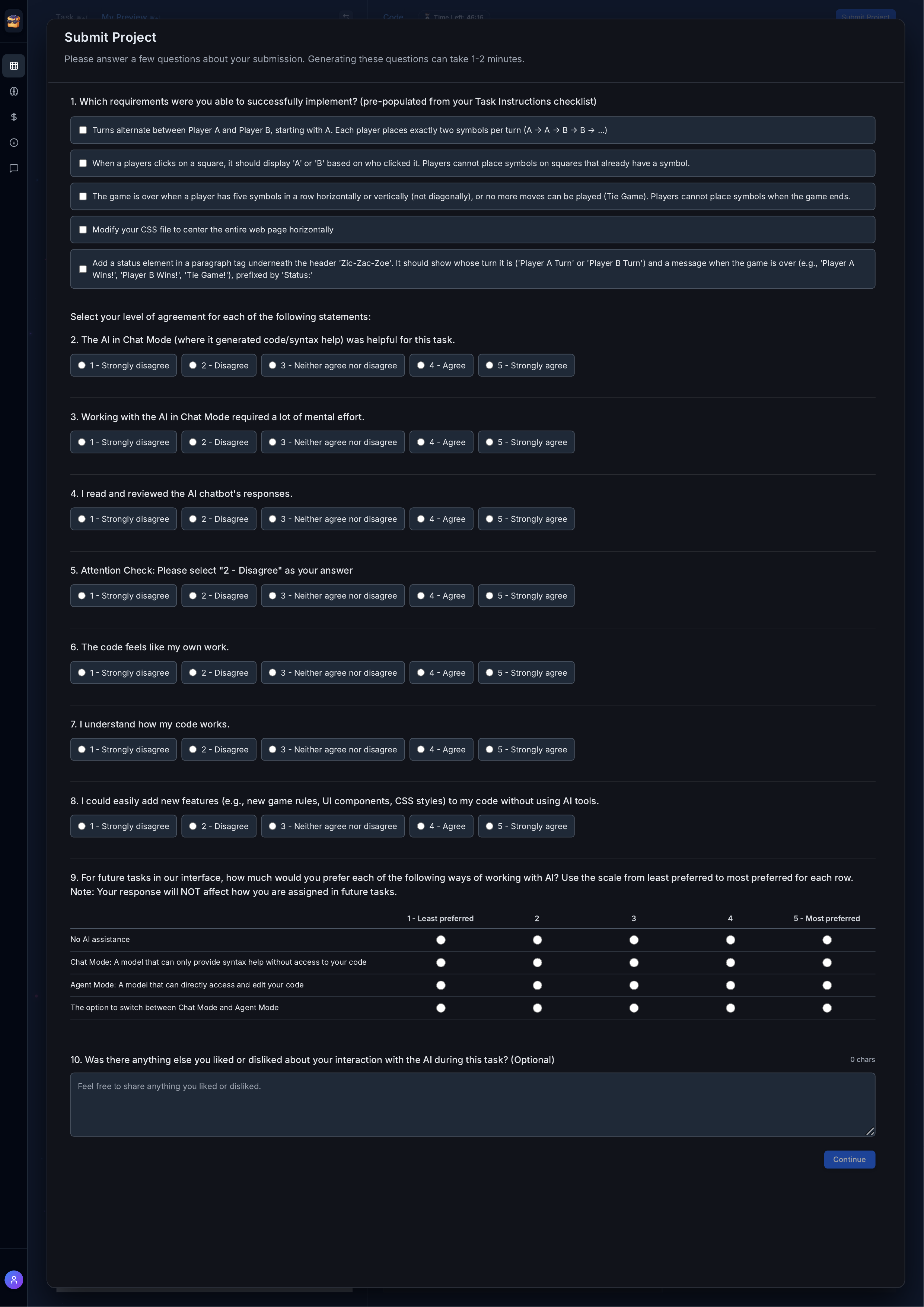}
    \caption{Example questions after the initial task submission shown to users in our interface for self-reporting usefulness, understanding, and preferences (\cref{subsection:preferences}).}
    \label{fig:ui:initial}
\end{figure*}

\begin{figure*}
    \centering
    \includegraphics[width=\linewidth]{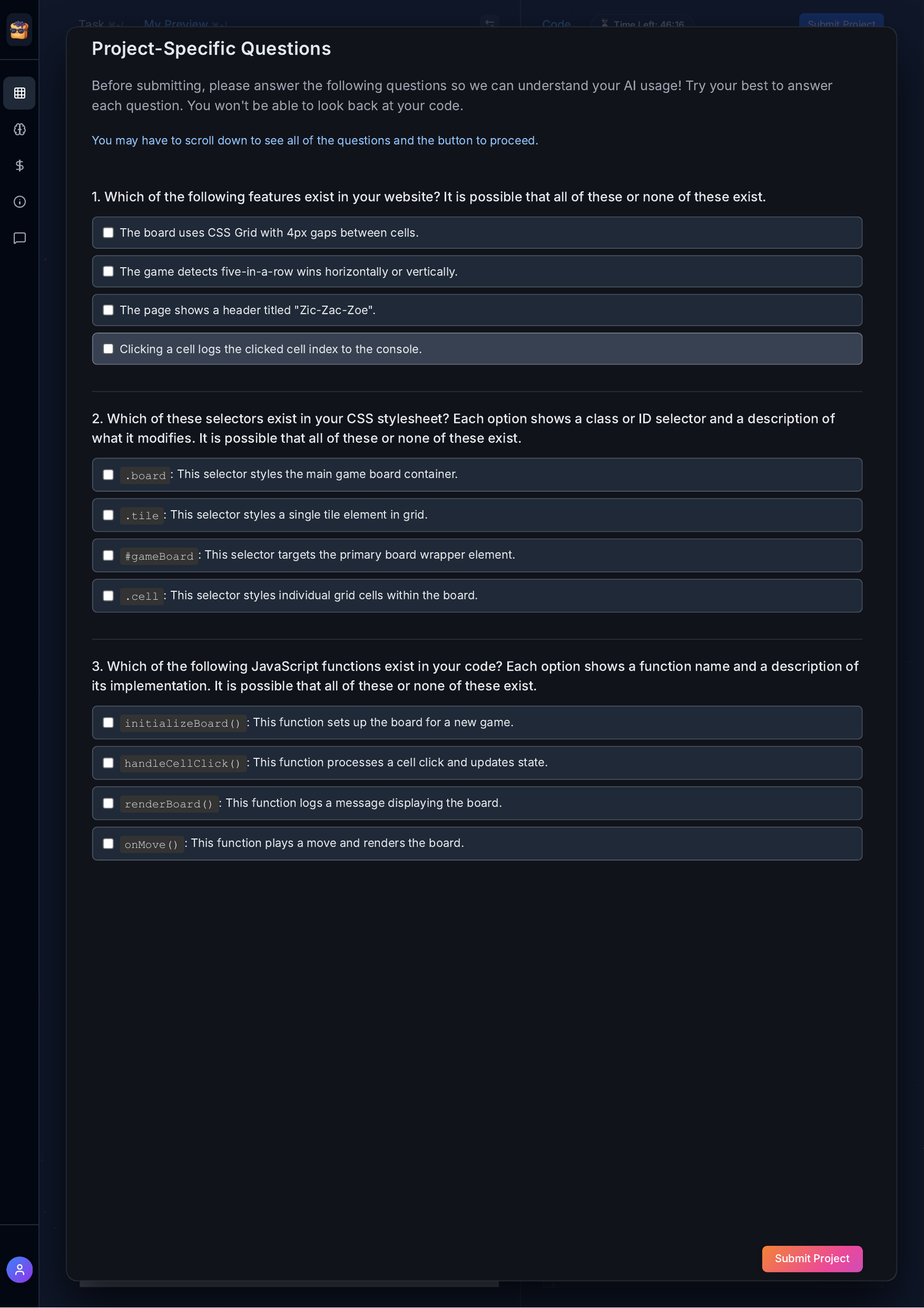}
    \caption{Example recall question in the comprehension assessment (\cref{subsection:comprehension}): selecting which functions, elements, and selectors exist in the users' code.}
    \label{fig:ui:recall_select}
\end{figure*}

\begin{figure*}
    \centering
    \includegraphics[width=\linewidth]{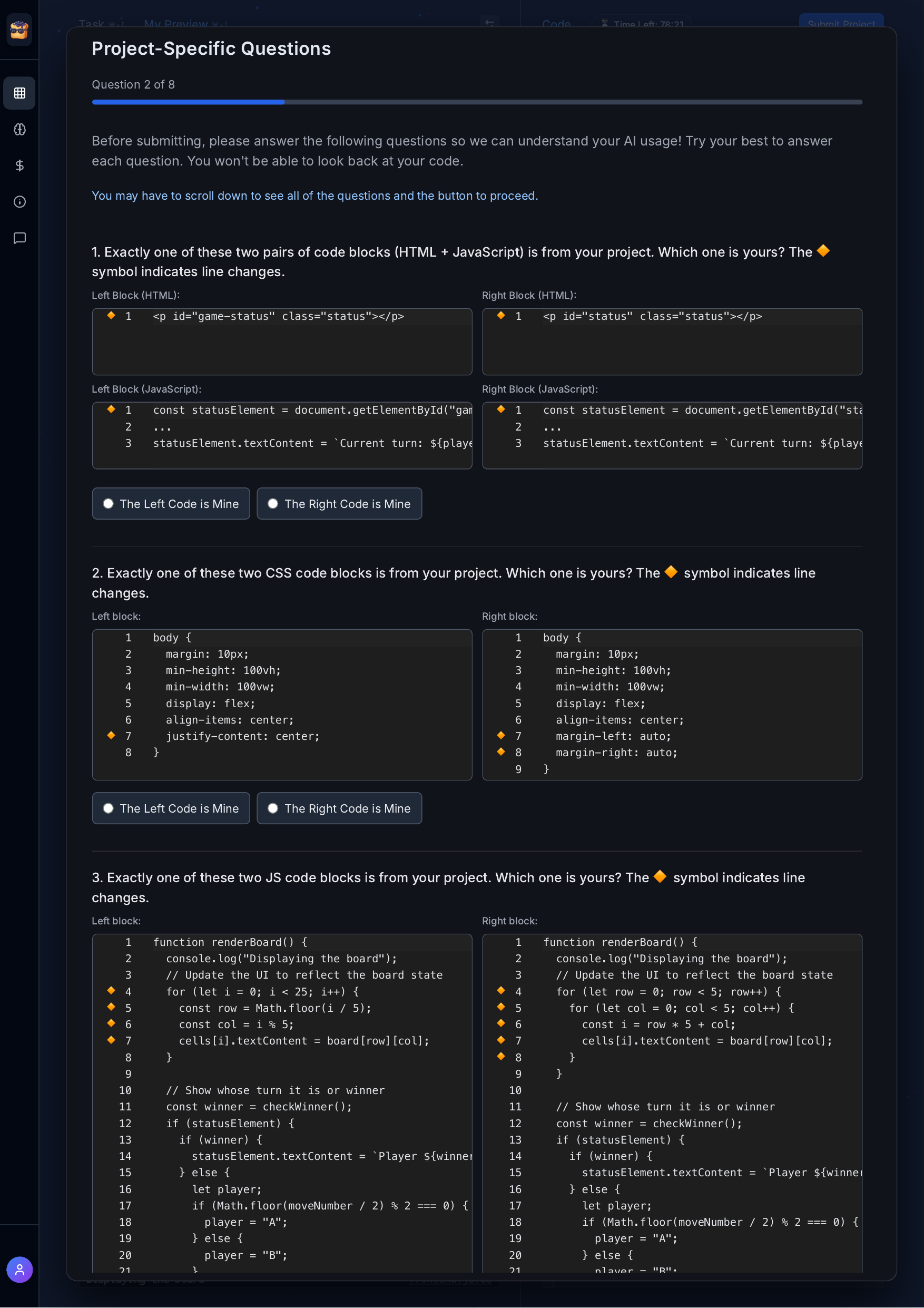}
    \caption{Example recall question in the comprehension assessment (\cref{subsection:comprehension}): selecting which of two logically-equivalent snippets the user created.}
    \label{fig:ui:recall_snippet}
\end{figure*}

\begin{figure*}
    \centering
    \includegraphics[width=\linewidth]{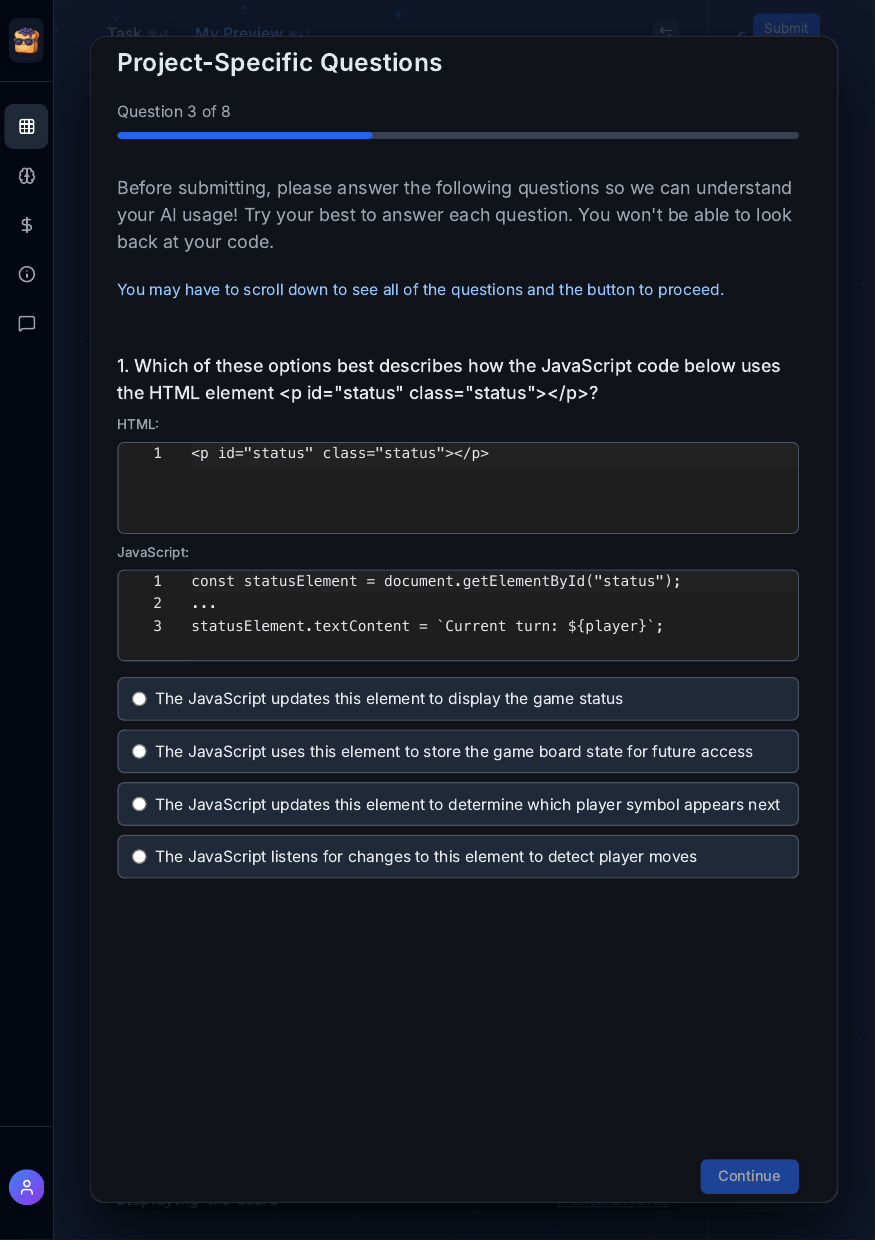}
    \caption{Example code reasoning in the comprehension assessment (\cref{subsection:comprehension}): identifying the purpose of a function in the users' code.}
    \label{fig:ui:reasoning}
\end{figure*}

\begin{figure*}
    \centering
    \includegraphics[width=\linewidth]{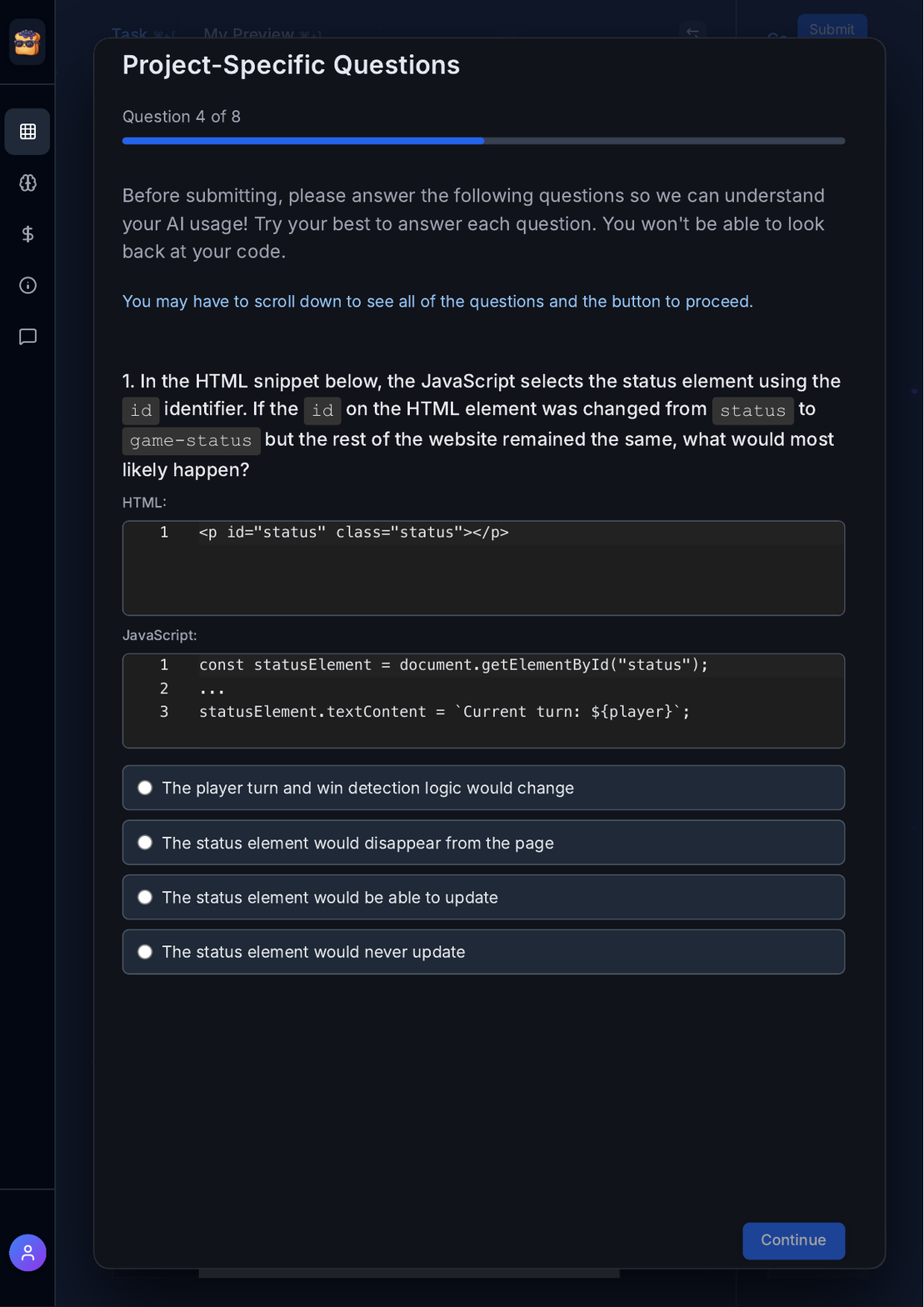}
    \caption{Example code reasoning in the comprehension assessment (\cref{subsection:comprehension}): identifying how the user's game would change if their code changed.}
    \label{fig:ui:reasoning_change}
\end{figure*}
\clearpage
\begin{trace}[title={Rubric \thetcbcounter: Initial Task}, label=rubric:initial]
- The first symbol placed in a new game is A (the game starts with Player A's turn). If there is no turn placing logic, this rule is automatically violated.\\
- Turns alternate between Player A and Player B. If there is no turn placing logic, this rule is automatically violated.\\
        - Each player places exactly two symbols per turn before the turn switches. If there is no turn placing logic, this rule is automatically violated.\\
        - Clicking an empty square places the correct symbol (A or B) based on whose turn it is. If there is no turn placing logic, this rule is automatically violated.\\
        - Players cannot place a symbol on a square that already contains a symbol. If there is no turn placing logic, this rule is automatically violated.\\
        - After a game ends, no additional symbols can be placed. If there is no win detection logic or no end game states can be reached, this rule is automatically violated"\\
        - The game correctly detects horizontal five-in-a-row win conditions. If there is no win detection logic, this rule is automatically violated.\\
        - The game correctly detects vertical five-in-a-row win conditions. If there is no win detection logic, this rule is automatically violated.\\
        - The game has no other win conditions (e.g., diagonal wins). If there is no win detection logic, this rule is automatically violated.",
        "If the board fills with no winner, the game ends in a tie. If there is no win detection logic, this rule is automatically violated.\\
        - A status element visually appears on the page prefixed by “Status:”. Minor variations of 'Status' are fine and should NOT be considered violations.\\
        - The status element correctly displays whose turn it is. If there is no turn changing logic, this rule is automatically violated.\\
        - The status element correctly displays the game outcome (Player A wins, Player B wins, or Tie Game). If there is no game detection logic, this rule is automatically violated.\\
        - The entire web page is centered horizontally. If it is also centered vertically, it is NOT considered a violation.\\
        - Horizontal centering of the page is implemented by modifying the CSS file. If the page is not centered horizontally, this rule is automatically violated.
\end{trace}

\begin{trace}[title={Rubric \thetcbcounter: Extension Task}, label=rubric:extension]
        - The placed “A” symbols have a red color. If there is no turn placing logic, this rule is automatically violated.\\
        - The placed “B” symbols have a blue color. If there is no turn placing logic, this rule is automatically violated.\\
        - The symbol colors are defined in the CSS file rather than set directly in JavaScript. If there is no turn placing logic, this rule is automatically violated.\\
        - Player A wins the game if they place the A symbol on all four corners of the board. If there is no win detection logic, this rule is automatically violated.\\
        - Player B wins the game if they place the B symbol on all four corners of the board. If there is no win detection logic, this rule is automatically violated.\\
        - Players can still win using the win conditions from the original submission. In the first task, the user attempted to implement win-checking by looking for horizontal and vertical symbols in a row, which may or may not be implemented correctly. This rule's passing or failure should not depend on the horizontal or vertical detection logic. Instead, you must ensure that any win detection logic that was not related to the four corners is still preserved. If there is no win detection logic, this rule is automatically violated.\\
        - A 'Reset Game' button appears visually below the board\\
        - Clicking the reset button clears the board of symbols and colors. If either the colors or symbols are not reset properly, this rule is automatically violated. Both conditions must be met for the rule to be passed\\
        - Clicking the reset button resets the game so it is Player A's turn. If there is no reset button, this rule is automatically violated.\\
        - Clicking the reset button resets the display of the status element. If there is no reset button, this rule is automatically violated.\\
        - The reset button does not refresh the page. If there is no reset button, this rule is automatically violated.\\
        - After reset, the game accepts moves and turn progression. If there is no reset button, this rule is automatically violated
\end{trace}
\hypersetup{
    linkcolor=white,
    citecolor=white,
    urlcolor=white
}

\lstset{
  literate={<}{{<}}1
           {>}{{>}}1
}

\begin{prompt}[title={Prompt \thetcbcounter: Chatbot Prompt}, label=prompt:chatbot]
You are an assistant that can answer syntax questions for HTML, CSS, and JavaScript code.\\

Rules:\\
1. You can only provide syntax-help guidance based on the user's existing code and errors.\\
2. You must not generate project/content-specific implementation code (for example: feature code, game logic, UI components, or task-completion code). You should judge the user's prompt and decide whether they are trying to bypass these safeguards.\\
3. If code is needed for syntax clarification, provide at most 3 lines total, and only as minimal syntax examples directly tied to syntax usage. If you can only comply with the user's request by generating more than three lines of code (e.g., for loop(s) with multiple lines of logic within the loop), that is a sign that this is not an appropriate syntax question.\\
4. If the user asks for content-specific code, the implementation of specific algorithms, the design of large HTML components, the creation of large CSS style sheets, asks you to complete parts of their project, or tries to bypass these constraints, politely refuse.\\
5. During refusal, explicitly state that you cannot edit their code and can only provide syntax guidance. While refusing, you do not provide the code block.\\
6. Do not claim to run commands or tools.\\
7. If the user pastes a code snippet or function body, refuse to debug it, refuse to point out errors, and refuse to suggest implementation details.\\
8. If the user asks about a specific error message, you can respond with what the error message means and what they could look for, but you should not provide the exact patch to fix it.\\
9. Never tell the user to switch to another mode.\\
10. Keep responses concise and actionable.\\
11. Do not repeat the phrase "For syntax guidance" excessively\\
12. If you are not directly addressing the user's input request due to refusal, you should acknowledge what you are not complying with and what you are going to do instead\\

Before responding to the user's request, think hard if you should refuse the user's request. Think about what an educator would do who does not want to just give the answer to the user. 
\end{prompt}

\begin{prompt}[title={Prompt \thetcbcounter: Summary Generation}, label=prompt:summary]
You are an expert at summarizing actions that an AI assistant took after being prompted by a user and providing useful suggestions for the user to improve their code.\\

This is what the user asked the assistant to do:\\
<query>\\
\texttt{[insert query]}\\
</query>\\

These are the final versions of files after edits (only changed files included):\\
<final\_files>\\
\texttt{[insert files]}\\
</final\_files>\\

These are the changes that the assistant made to the code (with optional SEARCH/REPLACE edit blocks when available):\\
<changes>\\
\texttt{[insert list of edits]}\\
</changes>\\

Using this information your job is to generate:\\
1. A summary of the changes that the assistant made to the code.\\
2. A list of ideas for the user to improve their code.\\

<summary instructions>\\
- The summary should be written in first person as if you were the one who made edits to the code. Use "I" as appropriate.\\
- You must discuss which files were edited and the specific changes to each file.\\
- Be subtle in how the changes address the user's request; do not quote the user's request.\\
- Be concise. The summary should be a maximum of two sentences.
</summary instructions>\\

<idea instructions>\\
- Generate 3 ideas with their corresponding probabilities, sampled from the full distribution.\\
- Each idea should improve the code or the task: e.g. task fulfillment, correctness, style, readability, edge cases, or user experience, depending on what fits the project (web UI, Python script, etc.).\\
- Only suggest ideas that are feasible given the file types and stack (e.g. for web: HTML/CSS/JS; for Python: standard library, tests, clarity). Do not suggest custom assets, external services, or out-of-scope changes.\\
- Frame each idea as a follow-up action the user could ask for, i.e. a short command starting with a verb.\\
- Be concise. Each idea should be no more than 10 words.\\
</idea instructions>\\

<format instructions>\\
Generate your output as a json with two keys: 1) "summary" with a string value of the summary; 2) "ideas" with a list of strings value of the ideas; and 3) "probabilities" with a list of floats value of the probabilities of each idea based on your full distribution.\\
\texttt{[insert JSON]}\\
Do not generate anything else\\
</format instructions>
\end{prompt}

\begin{prompt}[title={Prompt \thetcbcounter: Identifier Distractor Prompt}, label=prompt:distractor_identifier]
<task>
You are an expert at generating function names that do not exist in a user's code but plausibly could.\\

Given the existing function names and their implementations below, generate exactly four function names that mimic the style and domain of the existing code but do not actually exist. Use the implementations to understand what the code does (e.g. game turns, setup, UI) so your fake names are plausible. For example, if the code has "playTurn" and "setupLogic", you could generate "endTurn", "setupGame", "resetBoard". We will show these to users and ask them to identify which names exist and which do not, testing their comprehension of their own code.
</task>\\

Here are the existing functions (name and implementation):\\
<functions>
\texttt{[insert functions]}
</functions>\\

<function requirements>\\
- Mimic the style and domain of the existing function names; use the implementations to inform plausible fake names.\\
- None of the generated function names should be the same as the existing function names: \texttt{[insert function names]}. This is extremely important.\\
- You may generate: 1) names for features that do not exist in the code; 2) wrapper/helper names that do not exist; 3) names that suggest further decomposition of the real logic.\\
</function requirements>\\

<format>\\
Generate your output as a JSON with the key "fake\_function\_names" and the value being an array of exactly four function names as strings:\\
\texttt{[insert JSON]}

Do not generate anything else.
</format>
\end{prompt}

\begin{prompt}[title={Prompt \thetcbcounter: Snippet Distractor Prompt}, label=prompt:distractor_snippet]
<task>\\
Generate a distractor for this code block.\\
- Introduce one noticeable but logically equivalent difference such that someone who did not actually implement this code themselves may not realize the difference\\
- Do not add obvious stylistic differences or artifacts. For example, if the original code block did not have comments, you should not add comments to the distractor block. If the original code block did have comments, you should preserve them.
</task>\\

<code>
{code}
</code>\\

<html-priority-targets>\\
For HTML distractors, prioritize the following changes in this order:\\
- switch the label of the element's ID to something that this user could plausibly specify. this distractor should match the style of the user's\\
- switch the label of the element's class name to something that this user could plausibly specify. this distractor should match the style of the user's\\
- any other change that is noticeable but different\\
</html-priority-targets>\\

<task-specific-css-focus>\\
For zic\_zac\_zoe centering-related CSS, generate a different but logically equivalent way that the user could have centered their code horizontally, i.e., by altering different attributes than what the user's code did. Try to use one that is similar to what they did in style. For example, if they used justify-content: center, you could use align-items: center and switching the flexbox direction (and vice versa).\\

There are different ways to center the component, but the distractor should focus on one that appears simple. If you add too many extra lines compared to the original its a clear giveaway.\\

If this is not possible, introduce some other logically equivalent difference (but they should not be identical)\\
</task-specific-css-focus>\\

<task-specific-js-focus>\\
For zic\_zac\_zoe board rendering implementations, add an logically equivalent change related to indexing. Prefer one of the distractors in this order:\\
- If the user loops over a 2D matrix (e.g., the 2D 'board' based on rows and columns) and then converts it into a 1D matrix index (access 'cells' via 5*i+j), swap the approach: loop over the 1D list (cells) and map it into a 2D matrix indices (board). The vice versa also applies\\
- Make sure the style is EXACTLY the same between the two versions, including comments, indentation, and newlines. Try to use a more conventional style for both.\\
- If this 2D vs 1D change is not possible because the user has not implemented the function this way, introduce some other logically equivalent change of the function (but they should not be identical)\\
</task-specific-js-focus>\\

<format>
Return strict JSON only:
\texttt{[insert JSON]}
</format>
\end{prompt}

\begin{prompt}[title={Prompt \thetcbcounter: Question Personalization Prompt}, label=prompt:template]
<task>
You are validating whether a multiple-choice question can be asked about a user's code snippet. You MUST do two things:\\
1) Rewrite the question so it aligns with the provided question template and code details.\\
2) Validate the answer options, and adapt their wording only if needed for this participant's code.\\

Interpret the code literally. Do not infer intended behavior beyond what the code actually does or attempts to do. The generated question is meant to test the understanding of the user's own code.\\

Question-rewrite rules:\\
- Try to adhere to the style of the question\_template.\\
- Fill in concrete details from the participant's code (e.g. element IDs, function names, selectors) where the template has placeholders.\\
- Write a single clear question that stands on its own; the snippet will be shown right after it\\
- Make sure the question is extremely clear and easy to understand\\

Validation rule:\\
- The designated gold option must be correct for the user's actual code.\\
- Every distractor option must be incorrect for the user's actual code.\\
- The gold option must describe what the code ACTUALLY does, not the intended behavior\\

**Abstention policy.** A question should be selected when the participant's code contains a recognizable artifact related to the target feature. If no such artifact exists, return status = "abstain". Use the following guidelines:\\
- HTML questions: Select the question if the code includes a status element intended to display game status on the page (e.g., a <p>, <div>, or similar element that appears to hold status text) and JavaScript code that attempts to access that element (e.g., selectElementById, querySelector, ...)\\
- CSS questions: Select the question if the code includes a CSS rule that appears to attempt positioning or aligning the board or page layout (i.e., centering rules like align-items or justify-content). The implementation may be incomplete or incorrect. If the user did not change their code based on the starter implementation, then you MUST abstain.\\
- JavaScript questions: Select the question if the code includes a function or logic that renders or populate the game board on the page (e.g., iterating over the board state and updating DOM elements). The implementation may be incomplete or incorrect. If this function (initially called renderBoard()) does not exist or was not attempted (blank or has a print statement), abstain.\\
If none of these artifacts are present for the relevant question type, return status = "abstain".\\

Option-adaptation rules:\\
- Start from the provided sample gold answer and sample distractors.\\
- Keep the original wording when it already fits the participant's code.\\
- If wording does not fit, adapt it so one option is clearly correct and three are clearly incorrect for this participant's implementation.\\
- Avoid introducing stylistic cues: keep all four options similar in length (no more than five word differences), tone, specificity, and structure. In particular, all distractors must be very similar in length to each other (and the gold option should be similar too) so that length does not cue the answer. Do not add extra punctuation like parentheses or semi-colons on individual answers.\\
- None of the options should include words that relate to "change"; this is especially important in the counterfactual-style questions. We don't want to cue the user based on how their implementation current works. So instead of saying "Your board would stay centered" or this would "no longer" work, you could say "Your board would be centered"\\
- Remember: one option must stay correct and three must stay incorrect.\\
</task>\\

<question\_spec>
\texttt{[insert question template]}
</question\_spec>

<full\_website\_context>
\texttt{[insert website]}
</full\_website\_context>

<selected\_snippet>
\texttt{[insert snippet]}
</selected\_snippet>

The question template below provides an example question choices for a WORKING implementation. If the user's implementation is correct, you can adapt this question directly. If their implementation is incomplete or incorrect, you will have to adapt the options to make sure the gold answer is correct and the distractors are objectively wrong.\\
<provided\_option\_templates>\\
- sample\_gold\_answer: \texttt{[insert gold answer]}\\
- sample\_distractors:\\
  - \texttt{[insert distractor 1]}\\
</provided\_option\_templates>\\

Remember, if you are changing the gold answer, keep the word length differences minimal! (less than 5 words)\\

<format> Return strict JSON only. Also provide a short reason explaining why the question was selected or why the feature was absent in a "reason" key, and how you made any changes: 
\texttt{[insert JSON]} </format>
\end{prompt}

\begin{prompt}[title={Prompt \thetcbcounter: Judge Prompt}, label=prompt:judge]
You are an expert at evaluating whether user's code adheres to rules from a rubric used as grading criteria.\\

A user was asked to create a fictional game called "Zic-Zac-Zoe" using HTML, CSS, and JavaScript. They had to implement these specifications:\\
<specifications the user saw>
\texttt{[insert requirements]}
</specifications the user saw>\\

They finished the assignment and submitted the following code:\\
<code>
\texttt{[insert code]}
</code>\\

Now, as educators, we must evaluate their code to give them a final grade. We constructed a rubric from the zic-zac-zoe task, and now we will score each rule one at a time. Here is the current rule from the full rubric that you are assessing:\\
<rule>
\texttt{[insert single rubric rule]}
</rule>\\

Your task is to judge whether the user's submitted code **passes or violates this rule**. Here are some guidelines for evaluation:\\
<evaluation instructions>\\
- Grade the code like an educator. Do not allow the user to be clever and bypass the rules; they must make a "good-faith" effort to follow the rule as intended \\
- There is no partial credit. Either the rule is "passed" or "violated"\\
- Grade the website as if you were looking at it visually. For this reason, if the code has syntax errors that would stop the whole site or certain parts from rendering, then you should not consider the unrendered parts as properly implemented. For example, if the JavaScript code has a syntax error, then none of the logic would run and should not be evaluated. We know that the HTML, CSS, and JavaScript are correctly linked together so cross-file integration will not be an issue.\\
</evaluation instructions>\\

<format>\\
Return your output as valid JSON in the following format:\\
\texttt{[insert JSON]}\\
Do not include anything else.\\
</format>
\end{prompt}

\begin{prompt}[title={Prompt \thetcbcounter: Cluster Labeling Prompt}, label=prompt:label]
You are an expert at classifying text into its most relevant cluster \\

You will analyze user prompting strategies as they work with an AI assistant to develop a game of tic-tac-toe. After the user submits their prompt, the agent directly modifies the user's code to adhere to that request. Your job is to classify from a list of cluster names, which best applies to the input prompt.\\

These were the exact rubric items the user had to meet with their submission:\\
<rubric>
\texttt{[insert rubric]}
</rubric>\\

Here are your instructions:\\
<instructions>\\
- Choose the single best cluster for the action from the list of clusters. Do not invent new cluster titles.\\
- Prefer a specific existing cluster whenever possible.\\
- Use "Other" only when no listed cluster is a reasonable semantic fit.\\
- Ensure you reference the <rubric> when making your decision\\
- Keep explanation concise but concrete and evidence-based.\\
</instructions>\\

Here are the name of clusters\\
<cluster names>\\
1. copy: the user directly copies one or more of the rubric requirements verbatim or nearly verbatim\\
2. rephrased: the user rephrases one of the rubric requirements in their own words, but still using natural language. Prompts in other languages fall into this category\\
3. technical: the user mentions specific HTML, CSS, or JavaScript syntax (e.g., HTML tags, function names, or CSS selectors/identifiers), rather than using pure natural language text\\
4. exploratory: the user asks the AI to summarize the codebase or about its abilities\\
5. debugging: the users asks for the AI to refine or debug an initial mistake \\
</cluster names>\\

Here is the input prompt you must classify:\\
<prompt>
\texttt{[insert prompt]}
</prompt>\\

<format>\\
Return a JSON with two keys: 1) a string "cluster\_title" which is the name of the cluster you chose to classify; and 2) a string "explanation" which explains your decision\\
\texttt{[insert JSON]}\\
Return ONLY the JSON object without extra text
\end{prompt}

\begin{prompt}[title={Prompt \thetcbcounter: Cluster Chatbot Labeling Prompt}, label=prompt:label_chatbot]
You are an expert at classifying text into its most relevant cluster \\

You will analyze user prompting strategies as they work with an AI assistant to develop a game of tic-tac-toe. After the user submits their prompt, the chatbot responds to that request, but it only helps the user by returning high-level syntax guidance; it never executes tasks on the user's behalf. Your job is to classify from a list of cluster names, which best applies to the input prompt.\\

These were the exact rubric items the user had to meet with their submission:\\
<rubric>
\texttt{[insert rubric]}
</rubric>\\

Here are your instructions:\\
<instructions>\\
- Choose the single best cluster for the action from the list of clusters. Do not invent new cluster titles.\\
- Prefer a specific existing cluster whenever possible.\\
- Use "Other" only when no listed cluster is a reasonable semantic fit.\\
- Ensure you reference the <rubric> when making your decision\\
- Keep explanation concise but concrete and evidence-based.\\
</instructions>\\

Here are the name of clusters\\
<cluster names>\\
1. syntax\_help: the user asks about basic HTML, CSS, or JavaScript syntax, operators, methods, DOM APIs, selectors, attributes, or language mechanics\\
2. design\_help: the user asks how to structure the task, reason through the game rules, manage turns, check win/tie conditions, or decompose the problem without requesting a full solution\\
3. snippet: the user asks for a specific localized code fragment, function, condition, loop body, counter, or partial implementation step\\
4. debugging: the user reports an error, broken behavior, unexpected output, layout issue, or asks the AI to fix/refine something that is not working\\
5. clarification: the user asks whether a prior suggestion is correct, where code should go, what a term means in context, or how to adapt a previous answer\\
6. jailbreaking: the user asks for a complete implementation, tries to bypass the chatbot's limits, copies or reframes the whole assignment as a request for finished code, or uses pressure tactics to obtain prohibited help\\
7. urgency: the user expresses frustration, distress, profanity, dissatisfaction, time pressure, or emotional escalation; this can be used as a secondary label alongside one of the other categories\\
</cluster names>\\

Here is the input prompt you must classify:\\
<prompt>
\texttt{[insert prompt]}
</prompt>\\

<format>\\
Return a JSON with two keys: 1) a string "cluster\_title" which is the name of the cluster you chose to classify; and 2) a string "explanation" which explains your decision\\
\texttt{[insert JSON]}\\
Return ONLY the JSON object without extra text\\
</format>
\end{prompt}

\end{document}